%% file: main.tex
\documentclass[lettersize,journal]{IEEEtran}
\usepackage{amsmath,amsfonts}
\usepackage{algorithmic}
\usepackage{algorithm}
\usepackage{array}
\usepackage[caption=false,font=normalsize,labelfont=sf,textfont=sf]{subfig}
\usepackage{textcomp}
\usepackage{stfloats}
\usepackage{url}
\usepackage{verbatim}
\usepackage{graphicx}
\usepackage{cite}
\usepackage{booktabs}
\usepackage{xcolor}
\usepackage{colortbl}
\usepackage{multirow}
\usepackage{tabularx}
\usepackage{enumitem}
\newcommand{\eg}{\textit{e.g.}}
\newcommand{\ie}{\textit{i.e.}}

\usepackage[accsupp]{axessibility}
\usepackage[pagebackref,breaklinks,colorlinks,citecolor=blue]{hyperref} 

\usepackage[capitalize]{cleveref}

\begin{document}
\title{CurveStream: Boosting Streaming Video Understanding in MLLMs via Curvature-Aware Hierarchical Visual Memory Management}

\author{Chao Wang\textsuperscript{1*},
 Xudong Tan\textsuperscript{1*},
 Jianjian Cao\textsuperscript{1},
 Kangcong Li\textsuperscript{1},
and~Tao Chen\textsuperscript{1,2\dag}
\thanks{\textsuperscript{*}Contributed equally to this work.}%
\thanks{\textsuperscript{\dag}Corresponding author.}%
\thanks{\textsuperscript{1}Chao Wang, Xudong Tan, Jianjian Cao, Kangcong Li are with the College of Future Information Technology, Fudan University, Shanghai, China (e-mail: chaowang25@m.fudan.edu.cn).}%
\thanks{\textsuperscript{1,2}Tao Chen is with the College of Future Information Technology, Fudan University, Shanghai, China, and also with Shanghai Innovation Institute,
Shanghai, China (e-mail: eetchen@fudan.edu.cn).}%
}


\maketitle

\input{sec/0-abstract}


\input{sec/1-intro}
\input{sec/2-related}

\input{sec/3-methods}
\input{sec/4-exp}

\input{sec/5-conclusion}

\bibliographystyle{IEEEtran}
\bibliography{main}

\appendices
\input{sec/appendix}

\end{document}

%% file: sec/0-abstract.tex
\begin{abstract}
Multimodal Large Language Models have achieved significant success in offline video understanding, yet their application to streaming videos is severely limited by the linear explosion of visual tokens, which often leads to Out-of-Memory (OOM) errors or catastrophic forgetting. Existing visual retention and memory management methods typically rely on uniform sampling, low-level physical metrics, or passive cache eviction. However, these strategies often lack intrinsic semantic awareness, potentially disrupting contextual coherence and blurring transient yet critical semantic transitions. To address these limitations, we propose CurveStream, a training-free, curvature-aware hierarchical visual memory management framework. Our approach is motivated by the key observation that high-curvature regions along continuous feature trajectories closely align with critical global semantic transitions. Based on this geometric insight, CurveStream evaluates real-time semantic intensity via a Curvature Score and integrates an online K-Sigma dynamic threshold to adaptively route frames into clear and fuzzy memory states under a strict token budget. Evaluations across diverse temporal scales confirm that this lightweight framework, CurveStream, consistently yields absolute performance gains of over 10\% (e.g., 10.69\% on StreamingBench and 13.58\% on OVOBench) over respective baselines, establishing new state-of-the-art results for streaming video perception.The code will be released at https://github.com/streamingvideos/CurveStream.

\end{abstract}

%% file: sec/1-intro.tex
\section{Introduction}
\label{sec:intro}

While Multimodal Large Language Models (MLLMs) have achieved remarkable success in offline video understanding~\cite{bai2025qwen3, wang2024qwen2, Qwen2.5-VL,li2025videochat,zhang2024llavanext-video,lin2024video}, their application to streaming video scenarios is still hindered by fundamental bottlenecks. Streaming videos are theoretically infinite in length, inevitably leading to a linear explosion of visual tokens. Under stringent GPU memory constraints, models are highly susceptible to Out-of-Memory (OOM) errors or suffer from catastrophic forgetting caused by naive truncation strategies~\cite{xu2025streamingvlm}. Consequently, continuously and dynamically managing visual memory within a fixed memory budget emerges as the core challenge in achieving long-term streaming video understanding.

\begin{figure*}[tb]
  \centering
  \includegraphics[width=\linewidth]{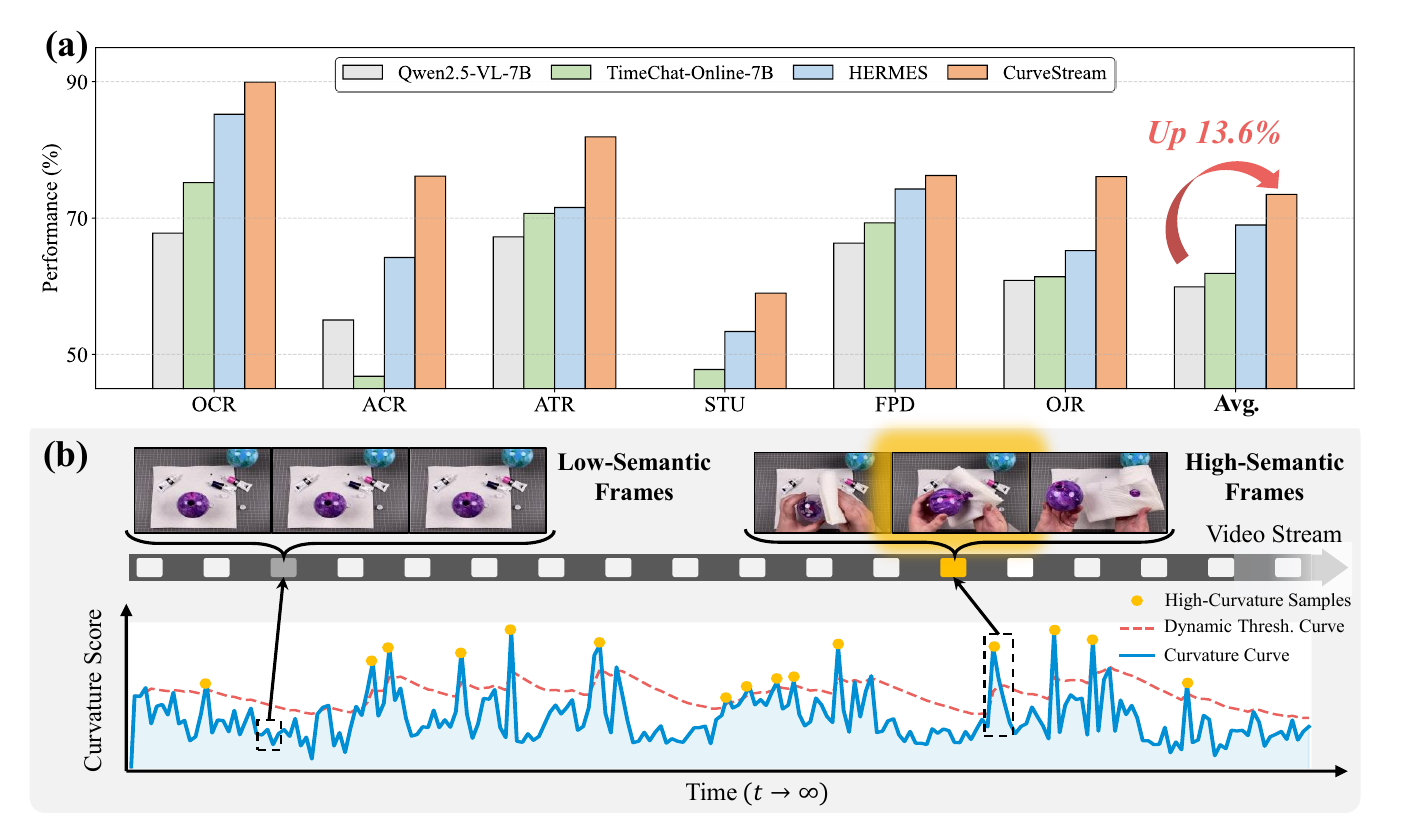}
\caption{\textbf{Performance and mechanism of CurveStream.} \textbf{(a)} CurveStream achieves state-of-the-art on OVOBench among training-free paradigms, boosting performance by 13.6\% over the Qwen2.5-VL-7B baseline. \textbf{(b)} Curvature-aware memory management over infinite streams ($t \rightarrow \infty$). By evaluating real-time semantic intensity (blue curve) against a K-Sigma dynamic threshold (pink dashed line), it adaptively filters redundant Low-Semantic Frames. Critical High-Semantic Frames (yellow dots) at curvature peaks are preserved, ensuring optimal visual context retention under strict token limits.}
  \label{fig:Performance_and_mechanism}
\end{figure*}

To address the challenge of linear token explosion, existing methods primarily focus on two aspects: visual information retention and long-term memory management. Visual information retention strategies typically utilize uniform sampling~\cite{tang2025adaptive,ye2025re,hu2025m} or low-level difference metrics (including inter-frame similarity~\cite{li2026freshmem,wang2025videollamb} or optical flow~\cite{xiong2025streaming}). However, these approaches are often sensitive to local noise and prioritize low-level physical motion, making it difficult to robustly capture the high-level global semantic transitions required for multimodal reasoning. Building upon these retained visual features, long-term memory management mechanisms further process the context. Mainstream solutions predominantly include rule-based cache eviction~\cite{zhang2026hermeskvcachehierarchical,kim2025infinipotv,xu2025streamingvlm,yang2025streammem}, feature clustering and merging, and retrieval paradigms utilizing external storage~\cite{di2025rekv}. 

Despite their progress, these visual retention and memory management methods share common limitations that hinder efficient streaming video understanding: 1) Semantic Fragmentation: They mostly employ passive eviction or smoothing compression strategies lacking intrinsic semantic awareness, which disrupts contextual coherence. 2) Information Blurring: During indiscriminate feature compression, they irreversibly blur transient yet critical semantic transition points. 3) Delayed Perception: Retrieval mechanisms conditioned on post-hoc queries restrict the model's capability for real-time, proactive perception in unbounded streaming scenarios.

To overcome these limitations, we re-examine the evolutionary dynamics of video streams within the feature space. We observe a critical phenomenon: when mapping a continuous video stream into a trajectory within the feature space, the high-curvature regions along this trajectory precisely correspond to high-quality visual semantic transitions.  Unlike uniform sampling or physical motion metrics that treat frames equally or focus on local noise, curvature geometrically measures the intensity of semantic shifts. A sharp turn (high curvature) in the feature trajectory signifies the emergence of a new event, a sudden viewpoint change, or a critical action boundary. This implies that utilizing ``curvature'' as an evaluation metric enables the precise extraction of the most valuable contextual information for reasoning, thereby offering a novel perspective for constructing highly efficient, adaptive streaming video memory management systems. As illustrated in Fig.~\ref{fig:Performance_and_mechanism} (b), this geometric approach effectively identifies critical semantic transitions by monitoring the trajectory's curvature peaks.

Building upon this curvature observation, we propose CurveStream, a training-free, curvature-aware hierarchical visual memory management framework. Diverging from uniform sampling strategies that periodically drop frames, we formulate streaming video processing as a dynamic, semantic-aware memory update process under a fixed token capacity limit ($N$). Specifically, CurveStream first calculates a Curvature Score in real time to represent the intensity of semantic transitions, integrating motion variation of consecutive frames with the geometric angle between feature displacement vectors. To achieve adaptive memory management in non-stationary video streams, we introduce an online-updating K-Sigma rule ($g = \mu + k\sigma$). This mechanism dynamically generates an admission threshold based on the running mean and variance of the historical curvature, adaptively categorizing high-value visual tokens into distinct hierarchical states (Clear Memory and Fuzzy Memory). When the memory bank reaches its capacity limit, the system systematically evicts the oldest tokens following strict queue rules. This design ensures that models maintain an acute perception of core visual semantic trajectories under a constant memory footprint.

To comprehensively evaluate CurveStream, we conduct extensive experiments across diverse temporal scales, encompassing 10 Real-Time Visual Understanding tasks in StreamingBench~\cite{lin2024streamingbench}, 6 Real-Time Visual Perception tasks in OVOBench~\cite{niu2025ovo}, and 3 offline video datasets (15–1200s)~\cite{li2024mvbench,mangalam2023egoschema,fu2025videomme}. As a lightweight, model-agnostic module, CurveStream demonstrates broad architectural compatibility across the LLaVA-OneVision and Qwen-VL (2/2.5/3) series at 4B, 7B, 8B, and 32B parameter scales. As shown in Fig.~\ref{fig:Performance_and_mechanism}a, integrating our framework into the Qwen2.5-VL-7B baseline yields accuracies of 84.00\% and 73.48\% on StreamingBench and OVOBench, respectively, delivering absolute performance gains of 10.69\% and 13.58\%. Furthermore, CurveStream enables 7B-parameter open-source models to consistently surpass closed-source commercial systems, including GPT-4o and Gemini 1.5 Pro, validating its robust generalizability and practical efficacy.

In summary, the main contributions of this paper are as follows:

\begin{enumerate}
    \item Revealing the ``curvature'' effect in streaming videos. We discover that high-curvature regions in the latent feature space align with critical global semantic transitions, providing a geometric metric for evaluating visual information that overcomes local noise.
    
    \item Proposing CurveStream, a training-free hierarchical memory management framework. By integrating real-time curvature scoring with a dynamic K-Sigma threshold, it adaptively routes frames into clear and fuzzy memory states to handle non-stationary streams under fixed token budgets.
    
    \item Achieving state-of-the-art performance on streaming benchmarks CurveStream effectively mitigates OOM issues and consistently improves diverse MLLMs by approximately 10\% in streaming scenarios, showing broad applicability on benchmarks like StreamingBench and OVOBench.
\end{enumerate}

%% file: sec/2-related.tex
\section{Related Work}
\label{sec:related_work}

\subsection{Existing Visual Information Retention Strategies}
\label{subsec:visual_retention}

Existing strategies for visual information retention in long videos encompass various directions, with prominent approaches focusing on rule-based token compression and query-driven feature retrieval~\cite{li2025videochat,li2024llava,tu2025favor,liu2023llavavideo}. Rule-based methods mitigate redundancy by evaluating local feature similarities. AKS~\cite{tang2025adaptiveaks} and M-LLM~\cite{hu2025mM-LLM} employ adaptive keyframe selection algorithms to maximize video coverage. FLoC~\cite{cho2025floc}, FlexSelect~\cite{luflexselect}, and METok~\cite{wang2025metok} dynamically prune redundant tokens during inference utilizing attention weights or facility location functions. Query-driven approaches perform goal-oriented extraction by fetching relevant frames conditioned on user instructions. DIG~\cite{di2025dig}, APVR~\cite{gao2025apvr}, BOLT~\cite{liu2025bolt}, and MemVid~\cite{tangdivid} compute semantic similarities between post-hoc text queries and visual frames. These paradigms generally rely on delayed user queries or low-level physical metrics (including inter-frame cosine similarity). This makes them susceptible to local motion noise in dynamic scenes and limits their capacity for proactive perception. To address this, our method diverges from traditional metrics by leveraging the ``curvature'' of feature trajectories in the feature space. This perspective intrinsically captures global semantic transitions, ensuring robust retention that is resilient to local physical disturbances.

\subsection{Existing Streaming Video Memory Management Mechanisms}
\label{subsec:memory_management}

Processing theoretically infinite streaming videos inherently causes a linear explosion in memory footprint. To circumvent this, current mechanisms explore various solutions, with KV cache eviction and external structured memory being widely adopted~\cite{di2025rekv,chen2025streamkv,zhang2025flash}. KV cache eviction strategies passively discard historical tokens. InfiniPot-V~\cite{kim2025infinipotv}, StreamingTOM~\cite{chen2025streamingtom}, StreamingVLM~\cite{xu2025streamingvlm}, and HERMES~\cite{zhang2026hermeskvcachehierarchical} utilize sliding windows or spatio-temporal redundancy metrics to evict older tokens upon reaching a memory threshold. External memory approaches offload long-term context to expand capacity. StreamForest~\cite{zeng2025streamforest}, ReKV~\cite{di2025rekv}, VideoLucy~\cite{zuo2025videolucy}, and Venus~\cite{ye2025venus} organize video segments into hierarchical trees or move features to external storage, utilizing retrieval mechanisms to reactivate necessary context. However, these mechanisms treat memory management as a queue-based smoothing process or an isolated retrieval task. Consequently, they may blur transient semantic shifts and disrupt natural in-context coherence. In contrast, we formulate memory management as a dynamic, semantic-aware, in-context update process. CurveStream incorporates an online K-Sigma rule to actively evaluate historical curvature, adaptively categorizing and replacing clear and fuzzy memory within a strict token limit.

%% file: sec/3-methods.tex
\begin{figure*}[tb]
  \centering
  \includegraphics[width=\linewidth]{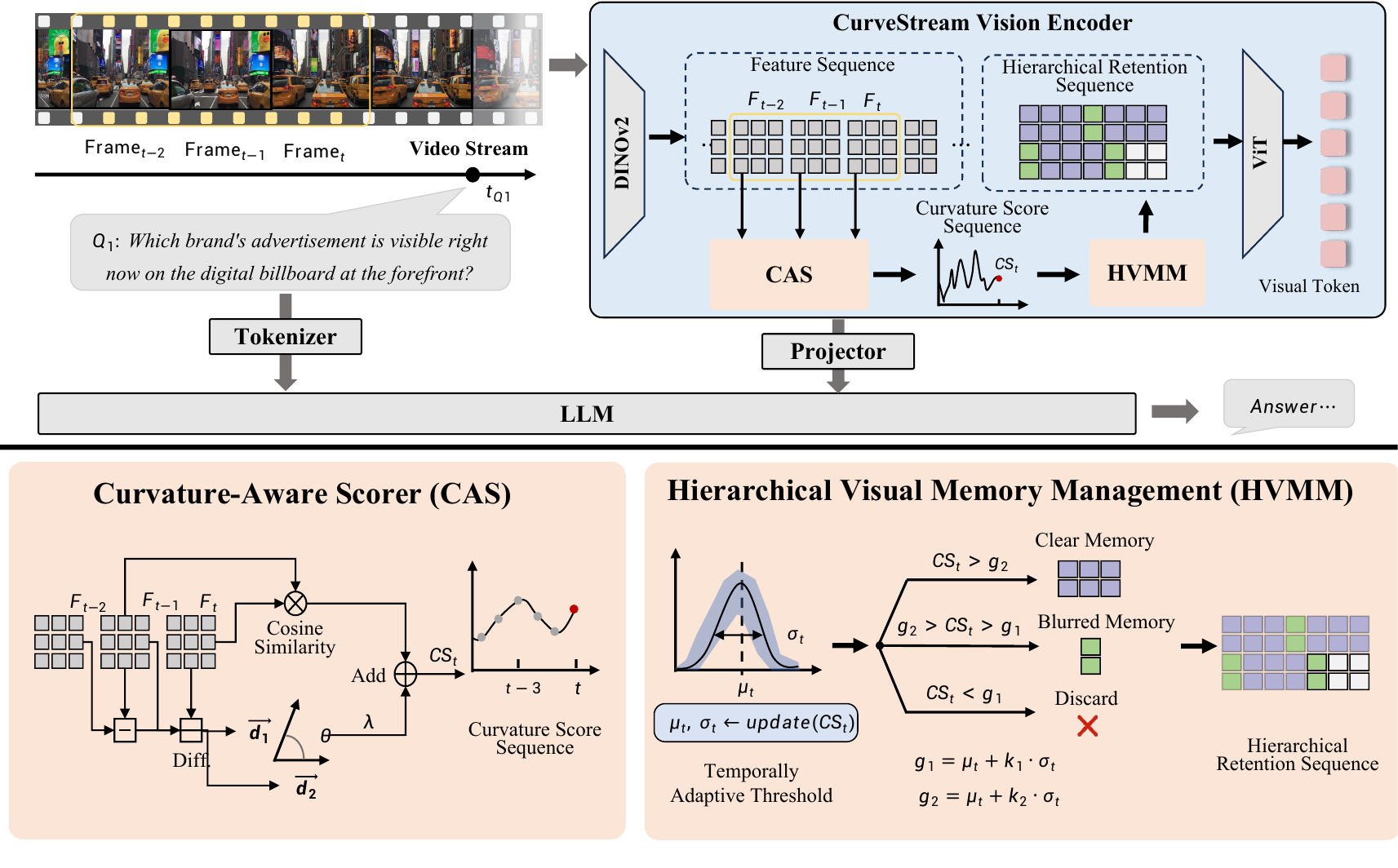}
  \caption{Overview of the CurveStream framework. This training-free vision encoder enables infinite streaming video understanding by replacing traditional sampling with a dynamic-retention perception layer designed to prevent Out-of-Memory (OOM) errors in long-term sequences. The Curvature-Aware Scorer (CAS) evaluates semantic transition intensity by fusing first-order motion variation and second-order trajectory curvature within the latent feature manifold, while the Hierarchical Visual Memory Management (HVMM) module dynamically routes incoming tokens into a fixed-capacity ($N_{\max}$) queue. By utilizing temporally adaptive $K$-Sigma thresholds, the encoder adaptively categorizes visual information into Clear, Blurred, or Discard states based on the intensity of semantic shifts, thereby ensuring a constant memory footprint while preserving critical visual anchors for long-term multimodal reasoning.
  }
  \label{fig:curvestream}
\end{figure*}

\section{Methods}
\label{sec:methods}

To achieve precise understanding of infinitely long streaming videos under strict memory constraints, we propose CurveStream, a training-free vision encoder architecture (illustrated in \cref{fig:curvestream}). The framework operates as an online selective-retention pipeline: it first utilizes a Curvature-Aware Scorer (CAS) to extract semantic transition intensity from the latent feature manifold trajectory, which is then processed by a Hierarchical Visual Memory Management (HVMM) module. Guided by temporally adaptive thresholds derived from online manifold statistics, this mechanism dynamically routes incoming frames into a fixed-capacity memory bank, categorizing them as Clear, Blurred, or Discarded.

\subsection{Problem Formulation}
\label{sec:formulation}

Let $\mathcal{V} = \{I_t\}_{t=1}^{\infty}$ be an infinitely long, continuous video stream, where $I_t$ denotes the visual observation at time step $t$. Suppose the system receives a natural language query $Q$ regarding the current or historical states at timestamp $t_q$. Due to the large parameter size $\Theta$ of Multimodal Large Language Models (MLLMs) and the quadratic complexity of self-attention mechanisms, it is computationally intractable to directly feed the entire historical sequence $\mathcal{V}_{\le t_q}$ into the model. Therefore, the system must maintain a dynamic visual memory queue $\mathcal{M}_t$ restricted by a maximum capacity limit $N_{\max}$.

We frame the streaming video understanding task as an online information extraction problem within a constrained space. At each time step $t$, the system needs to derive an efficient memory scheduling policy $\pi$. This policy evaluates the informative value of the current frame $I_t$ and outputs a state tuple $(s_{t},r_{t})$ containing the retention and resolution decisions to update the memory bank:
\begin{equation}
    \mathcal{M}_t = \text{Update}(\mathcal{M}_{t-1}, I_t, s_t, r_t)
\end{equation}
where $s_t \in \{\text{Clear}, \text{Blurred}, \text{Discard}\}$ represents the hierarchical routing state, and $r_t$ denotes the corresponding spatial resolution. 

The primary optimization objective of CurveStream is to maximize the conditional probability of the MLLM generating the correct answer $A$ under a strict queue length constraint ($|\mathcal{M}_t| \le N_{\max}$):
\begin{equation}
    \max_{\pi} P(A \mid Q, \mathcal{M}_{t_q}; \Theta)
\end{equation}
To solve this online decision-making problem lacking direct supervisory signals, we leverage the intrinsic geometric properties of the visual feature manifold to construct a lightweight scheduling policy $\pi$, realized through the CAS and HVMM modules described below.

\subsection{Curvature-Aware Scorer (CAS)}
\label{sec:cas}

In continuous visual streams, adjacent frames often exhibit high temporal redundancy. Especially in embodied AI or first-person perspectives, traditional sampling strategies based on simple feature differences are highly prone to overfitting to large translational motions. To accurately localize high-value information, we design the Curvature-Aware Scorer (CAS).

CAS utilizes a frozen visual encoder to extract the global feature representation $\mathbf{F}_t \in \mathbb{R}^D$ of the input frame $I_t$, followed by $L_2$ normalization. To characterize the evolutionary trajectory of features within the latent space manifold, we integrate both the first-order motion intensity and the second-order geometric curvature. Based on the cosine similarity between consecutive frames, the first-order Motion Variation is defined as:
\begin{equation}
    M_t = 1 - \frac{\langle \mathbf{F}_t, \mathbf{F}_{t-1} \rangle}{\|\mathbf{F}_t\| \|\mathbf{F}_{t-1}\|}
\end{equation}

To filter out constant-velocity background changes caused by smooth camera movements, we compute an approximation of the second-order partial derivative of the feature trajectory. Let the feature displacement vectors of adjacent time steps be $\mathbf{d}_1 = \mathbf{F}_{t-1} - \mathbf{F}_{t-2}$ and $\mathbf{d}_2 = \mathbf{F}_t - \mathbf{F}_{t-1}$. The local Geometric Curvature of the feature manifold is approximately represented by the angular deviation between these displacement vectors:
\begin{equation}
    C_t = 1 - \frac{\langle \mathbf{d}_1, \mathbf{d}_2 \rangle}{\|\mathbf{d}_1\| \|\mathbf{d}_2\|}
\end{equation}
When $\mathbf{d}_1$ and $\mathbf{d}_2$ are aligned in direction, $C_t$ approaches $0$, indicating a smooth transition period. Conversely, when the direction of feature evolution changes abruptly (e.g., a new entity intrudes or a sharp viewpoint shift occurs), $C_t$ increases significantly. The final Curvature Score $\mathit{CS}_t$ is formulated as a linear combination of the two:
\begin{equation}
    \mathit{CS}_t = M_t + \lambda C_t
\end{equation}
where $\lambda$ serves as the balancing coefficient for the geometric penalty term.

\subsection{Hierarchical Visual Memory Management (HVMM)}
\label{sec:hvmm}

After obtaining the $\mathit{CS}_t$ sequence, the Hierarchical Visual Memory Management (HVMM) module utilizes temporally adaptive dynamic thresholds to route high-value frames into a fixed-capacity memory bank at differentiated resolution levels, effectively suppressing KV Cache bloat.

\subsubsection{Online Manifold Distribution Estimation} 
In untrimmed embodied or first-person streaming videos, the temporal pacing typically exhibits significant dynamics. For instance, a subject might suddenly break into a vigorous run after a prolonged period of stationary observation. Under such complex scenarios, employing any a priori static threshold is highly likely to lead to memory bank collapse or severe loss of critical information.

Therefore, HVMM models the filtering of high-value information as an online distribution-aware process. To capture the dynamic pacing of the video stream in real time, we update the transient expectation $\mu_t$ and variance $\sigma_t^2$ of the curvature scores using an Exponential Moving Average (EMA) formulation:
\begin{equation}
    \mu_t = \gamma \mu_{t-1} + (1 - \gamma) \mathit{CS}_t, \quad \sigma_t^2 = \gamma \sigma_{t-1}^2 + (1 - \gamma) (\mathit{CS}_t - \mu_t)^2
\end{equation}
where $\gamma \in (0, 1)$ is the momentum factor controlling the size of the historical observation window. As the time step $t$ advances, the newly observed curvature score $\mathit{CS}_t$ smoothly calibrates the transient distribution parameters in a recursive manner. Based on this online evolutionary mechanism, we construct Gaussian distribution-aware dynamic dual thresholds: $g_1 = \mu_t + k_1 \sigma_t$ and $g_2 = \mu_t + k_2 \sigma_t$ ($k_1 < k_2$). This design enables CurveStream to adaptively scale its sensitivity to visual shifts according to the current intensity of the scene.

\subsubsection{Hierarchical State Transition} 
Guided by the adaptive dual thresholds, HVMM executes a resolution-aware hierarchical state transition strategy. Specifically, the retention state $s_t$ for an incoming frame $I_t$ is dynamically determined as follows:
\begin{equation}
    s_t = 
    \begin{cases} 
      \text{Clear Memory}, & \text{if } \mathit{CS}_t \ge g_2 \\
      \text{Blurred Memory}, & \text{if } g_1 \le \mathit{CS}_t < g_2 \\
      \text{Discard}, & \text{if } \mathit{CS}_t < g_1 
    \end{cases}
\end{equation}

\textbf{Clear Memory.} Frames satisfying $\mathit{CS}_t \ge g_2$ break through the current local dynamic distribution and capture significant semantic shifts. The system retains their original high-resolution features ($r_t = \text{High}$) and stores them in the memory bank to support subsequent fine-grained visual reasoning. Notably, the current frame $I_{t_q}$ that triggers the query is deterministically assigned this state to ensure immediate context awareness.

\textbf{Blurred Memory.} Frames falling within $g_1 \le \mathit{CS}_t < g_2$ are identified as intermediate transitional observations consistent with the current dynamic pacing. To preserve necessary temporal causal associations and action coherence while significantly compressing token overhead, these frames are downsampled to a minimal resolution ($r_t = \text{Low}$) before storage.

\textbf{Discard.} Frames with $\mathit{CS}_t < g_1$ represent low-information redundant observations below the local expected mean. The system directly discards these features to protect the scarce memory space.

Finally, to ensure a constant memory footprint without OOM risks, whenever the memory bank exceeds its capacity $(|\mathcal{M}_{t}|>N_{max})$, the system executes a strict First-In-First-Out (FIFO) eviction, removing the oldest tokens from the queue regardless of their retention states.

%% file: sec/4-exp.tex
\section{Experiments}
\label{subsec:exp_setup}

\subsection{Experimental Setup}

\subsubsection{Datasets.} 
To comprehensively evaluate the effectiveness of the proposed adaptive visual memory framework under various temporal dynamics, we conducted extensive experiments across five mainstream multimodal benchmarks encompassing three video paradigms. As the core of our evaluation for streaming video understanding, we selected StreamingBench~\cite{lin2024streamingbench} and OVOBench~\cite{niu2025ovo}. These two benchmarks rigorously test the model's capability for long-range event association and instantaneous dynamic response within continuous data streams. To address complex dynamic scenes, we utilized EgoSchema~\cite{mangalam2023egoschema}, a highly challenging egocentric benchmark that rigorously tests the model's ability to accurately capture micro-actions and perform causal reasoning amidst drastic viewpoint changes and redundant backgrounds. Furthermore, to explore the extreme limits of memory capacity, we introduced VideoMME~\cite{fu2025videomme}, comprehensively examining the model's feature retention and generalizability across short, medium, and extremely long (up to several hours) contexts. Finally, we incorporated the MVBench~\cite{li2024mvbench} short video benchmark to verify that the system's dynamic frame filtering and resolution reduction strategies do not compromise the model's spatio-temporal perception of fine-grained local actions.

\subsubsection{Baselines.} 
Our comparative analysis involves two major categories of baseline methods. The first category comprises state-of-the-art open-source Multimodal Large Language Models (Base MLLMs), specifically including LLaVA-OneVision~\cite{li2024llava} and multiple iterations of the Qwen-VL series (i.e., Qwen2-VL~\cite{wang2024qwen2}, Qwen2.5-VL~\cite{Qwen2.5-VL}, Qwen3-VL~\cite{bai2025qwen3}). The second category encompasses recent advanced frameworks specifically optimized for streaming video understanding or long-context visual processing (SOTA Streaming Methods), including Flash-VStream~\cite{zhang2025flash}, FreshMem~\cite{li2026freshmem}, HERMES~\cite{zhang2026hermeskvcachehierarchical}, and ReKV~\cite{di2025rekv}. By integrating our proposed training-free memory module into the base MLLMs, we conduct a direct performance comparison with these specialized SOTA methods under strictly equivalent visual token constraints.

\subsubsection{Implementation Details.} 
In all comparative experiments, to ensure evaluation fairness and strictly simulate the physical GPU memory constraints inherent in streaming video processing, we establish a uniform memory bank capacity upper limit (i.e., a maximum token budget $N$) across all methods. At the feature extraction frontend of our framework, we employ the lightweight DINOv2-small model to acquire local geometric representations of temporal features. During the adaptive memory allocation phase, for high-curvature core transition frames that trigger clear memory, the system retains the native dynamic high-resolution input of the base model. Conversely, for blurred memory frames representing smooth transition states, the resolution is uniformly downsampled to a fixed $224 \times 224$ to conserve memory space. All benchmark evaluations are independently executed on a single inference GPU to fully validate the robustness of our framework under severely limited memory conditions.

\subsection{Online Benchmark Results}
\label{subsec:main_results}

\cref{tab:main_results} presents the quantitative evaluation results of various methods on the streaming video benchmarks. Under strict visual token capacity constraints, our method achieves stable and significant performance leaps across different base models. Specifically, when utilizing Qwen2-VL-7B as the base model, our method achieves accuracies of 81.04\% and 70.73\% on StreamingBench and OVO-Bench, respectively, yielding absolute performance gains of 12.0\% and 10.08\% compared to the uniform sampling baseline.

More importantly, among training-free streaming video understanding frameworks, our method establishes a new state-of-the-art (SOTA). Compared to recent advanced specialized streaming video methods (\eg, FreshMem and HERMES), our framework further achieves absolute accuracy improvements of 6.84\% and 4.06\% on StreamingBench and OVO-Bench, respectively. 

This comprehensively leading performance is directly attributed to our adaptive visual memory mechanism. By introducing manifold curvature as a dynamic prior, the framework not only effectively strips away redundant static backgrounds in long videos but also precisely allocates limited memory resources to high-frequency visual transition points. This strategy, which highly aligns the memory queue with the underlying dynamic evolution of the video, fundamentally overcomes catastrophic forgetting in long-range reasoning, thereby preserving the highest-quality temporal context for the model.

\begin{table*}[t!]
  \centering
    \caption{Quantitative comparison of average accuracy across 10 Real-Time Visual Understanding sub-tasks in StreamingBench and 6 Real-Time Visual Perception sub-tasks in OVOBench. Best results are highlighted in bold, with absolute gains over respective baselines in red. Our frame count (10-20) reflects the dynamically changing size of the adaptive memory queue.}
  \label{tab:main_results}
  \setlength{\tabcolsep}{24pt} 
  \begin{tabular}{l c c c}
    \toprule
    Method & Frame & StreamingBench~\cite{lin2024streamingbench} & OVOBench~\cite{niu2025ovo} \\
    \midrule
    Human & - & 91.46 & 93.20 \\
    \midrule
    
    \rowcolor{gray!15} \multicolumn{4}{l}{\textit{Proprietary MLLMs}} \\
    \midrule
    Gemini 1.5 Pro~\cite{gemini} & 1fps & 75.69 & 69.32 \\
    GPT-4o~\cite{hurst2024gpt4o} & 64 & 73.28 & 64.46 \\
    
    \rowcolor{gray!15} \multicolumn{4}{l}{\textit{Open-source Offline MLLMs}} \\
    \midrule
    Qwen2-VL-7B~\cite{wang2024qwen2} & 64 & 69.04 & 60.65 \\
    InternVL-V2-8B~\cite{chen2024internvl} & 16 & 63.72 & 60.73 \\
    \midrule
    
    \rowcolor{gray!15} \multicolumn{4}{l}{\textit{Open-Source Online MLLMs}} \\
    \midrule
    Flash-VStream-7B~\cite{zhang2025flash} & - & 23.23 & 29.86 \\
    VideoLLM-online-8B~\cite{chen2024videollmonline} & 2fps & 35.99 & 20.79 \\
    Dispider-7B~\cite{qian2025dispider} & 1fps & 67.63 & 54.55 \\
    TimeChat-Online-7B~\cite{yao2025timechat} & 1fps & 75.36 & 61.90 \\
    StreamForest-7B~\cite{zeng2025streamforest} & 1fps & 77.26 & 61.20 \\
    \midrule
    
    \rowcolor{gray!15} \multicolumn{4}{l}{\textit{Training-free Offline-to-Online Methods}} \\
    \midrule
    LLaVA-OneVision-7B~\cite{li2024llava} & 64 & 71.34 & 63.06 \\
    \quad + ReKV~\cite{di2025rekv} & 0.5fps & 69.22 & 57.33 \\
    \quad + HERMES~\cite{zhang2026hermeskvcachehierarchical} & 1fps & 73.23 & 66.34 \\
    \rowcolor{blue!5} \quad \textbf{+ Ours} & 10-20 & \textbf{75.12} (\textcolor{red}{$\uparrow$ 3.78}) & \textbf{70.57} (\textcolor{red}{$\uparrow$ 7.51}) \\
    \midrule
    
    Qwen2-VL-7B~\cite{wang2024qwen2} & 1fps & 69.04 & 60.65 \\
    \quad + HERMES~\cite{zhang2026hermeskvcachehierarchical} & 1fps & - & - \\
    \quad + Freshmem~\cite{li2026freshmem} & 1fps & 74.20 & 66.67 \\
    \rowcolor{blue!5} \quad \textbf{+ Ours} & 10-20 & \textbf{81.04} (\textcolor{red}{$\uparrow$ 12.00}) & \textbf{70.73} (\textcolor{red}{$\uparrow$ 10.08}) \\
    \midrule
    
    Qwen2.5-VL-7B~\cite{Qwen2.5-VL} & 1fps & 73.31 & 59.90 \\
    \quad + HERMES~\cite{zhang2026hermeskvcachehierarchical} & 1fps & 79.44 & 68.98 \\
    \rowcolor{blue!5} \quad \textbf{+ Ours} & 10-20 & \textbf{84.00} (\textcolor{red}{$\uparrow$ 10.69}) & \textbf{73.48} (\textcolor{red}{$\uparrow$ 13.58}) \\
    \midrule
    
    Qwen3-VL-8B~\cite{bai2025qwen3} & 1fps & 73.20 & 70.1 \\
    \rowcolor{blue!5} \quad \textbf{+ Ours} & 10-20 & \textbf{85.56} (\textcolor{red}{$\uparrow$ 12.36}) & \textbf{80.76} (\textcolor{red}{$\uparrow$ 10.66}) \\
    \bottomrule
  \end{tabular}
\end{table*}

\subsection{Offline Benchmark Results}
\label{subsec:generalization}

\cref{tab:longvideoresults} presents the evaluation results of our framework on the short-video benchmark (MVBench) and long-video benchmark (VideoMME). Although our adaptive memory framework is specifically designed for streaming video scenarios, it also exhibits strong generalization ability in conventional offline short- and long-video understanding tasks.

As can be observed, our method consistently brings stable performance improvements across different base models. For instance, when built upon Qwen2.5-VL-7B, our method achieves a 1.03\% absolute gain (up to 66.03\%) on MVBench, a fine-grained action-oriented short-video benchmark, compared with the uniform sampling baseline. Meanwhile, when integrated into LLaVA-OneVision-7B, our method also yields a 1.77\% absolute improvement (up to 59.44\%) on VideoMME, a comprehensive long-video benchmark.

It is worth noting that for Qwen2.5-VL-7B on VideoMME, there is a slight performance drop (from 64.52\% to 62.97\%). This is because, to maintain a strictly constant memory footprint without OOM risks over hours-long videos, the system inevitably trades off some fine-grained global details to preserve the most critical semantic transitions. These quantitative results sufficiently verify the universality and effectiveness of the proposed framework in offline settings.

\begin{table*}[t!]
  \centering
  \caption{Quantitative comparison of the average accuracy on MVBench (20 sub-tasks), EgoSchema, and VideoMME benchmarks. The best results are highlighted in bold, with absolute performance gains over the respective baselines indicated in red.}
  \label{tab:longvideoresults}
  \resizebox{0.9\textwidth}{!}{
  \begin{tabular}{@{} l c c c c @{}}
    \toprule
    Method & Frame & MVBench~\cite{li2024mvbench} & EgoSchema~\cite{mangalam2023egoschema} & VideoMME~\cite{fu2025videomme} \\
    \midrule
    
    \rowcolor{gray!15} \multicolumn{5}{l}{\textit{Proprietary MLLMs}} \\
    \midrule
    GPT4-V~\cite{zhang2023gpt4v}  & 1fps & 43.7 & 55.6 & 60.7 \\
    GPT-4o~\cite{hurst2024gpt4o} & 64 & 64.6 & 72.2 & 77.2 \\
    \midrule
    
    \rowcolor{gray!15} \multicolumn{5}{l}{\textit{Open-source Offline MLLMs}} \\
    \midrule
    LLaVA-NeXT-Video~\cite{zhang2024llavanext-video} & 32 & 33.7 & 43.9 & 46.5 \\
    Qwen2-VL-7B~\cite{wang2024qwen2} & 64 & 67.0 & 66.70 & 69.0 \\
    VideoChat2~\cite{li2024mvbench} & 16 & 60.4 & 54.4 & 54.6 \\
    VideoLLaMA2~\cite{damonlpsg2024videollama2} & 32 & 54.6 & 51.7 & 46.6 \\
    \midrule
    
    \rowcolor{gray!15} \multicolumn{5}{l}{\textit{Open-Source Online MLLMs}} \\
    \midrule
    Dispider-7B~\cite{qian2025dispider} & 1fps & - & 55.60 & 57.20 \\
    \mbox{TimeChat-Online-7B~\cite{yao2025timechat}} & 1fps & 75.36 & 61.90 & 53.22 \\
    StreamForest-7B~\cite{zeng2025streamforest} & 1fps & 70.20 & - & 61.40 \\
    \midrule
    
    \rowcolor{gray!15} \multicolumn{5}{l}{\textit{Training-free Offline-to-Online Methods}} \\
    \midrule
    Qwen2.5-VL-7B~\cite{Qwen2.5-VL} & 1fps & 65.00 & 58.47 & \textbf{64.52} \\
    \quad + HERMES~\cite{zhang2026hermeskvcachehierarchical} & \text{1fps} & 65.53 & 59.47 & 60.63 \\
    \rowcolor{blue!5} \rule{0pt}{1.2em} \quad \textbf{+ Ours} & 1fps & \hphantom{($\uparrow$ 0)}\textbf{66.03}(\textcolor{red}{$\uparrow$ 1.03}) & \hphantom{($\uparrow$ 5.82)}\textbf{64.29}(\textcolor{red}{$\uparrow$ 5.82}) & 62.97 \\
    \bottomrule
  \end{tabular}
  }
\end{table*}

\subsection{Scalability Across Model Parameters}
\label{subsec:scalability}

\begin{figure*}[!t]
\centering
\subfloat[Scaling Performance Comparison]{%
    \includegraphics[width=0.48\textwidth]{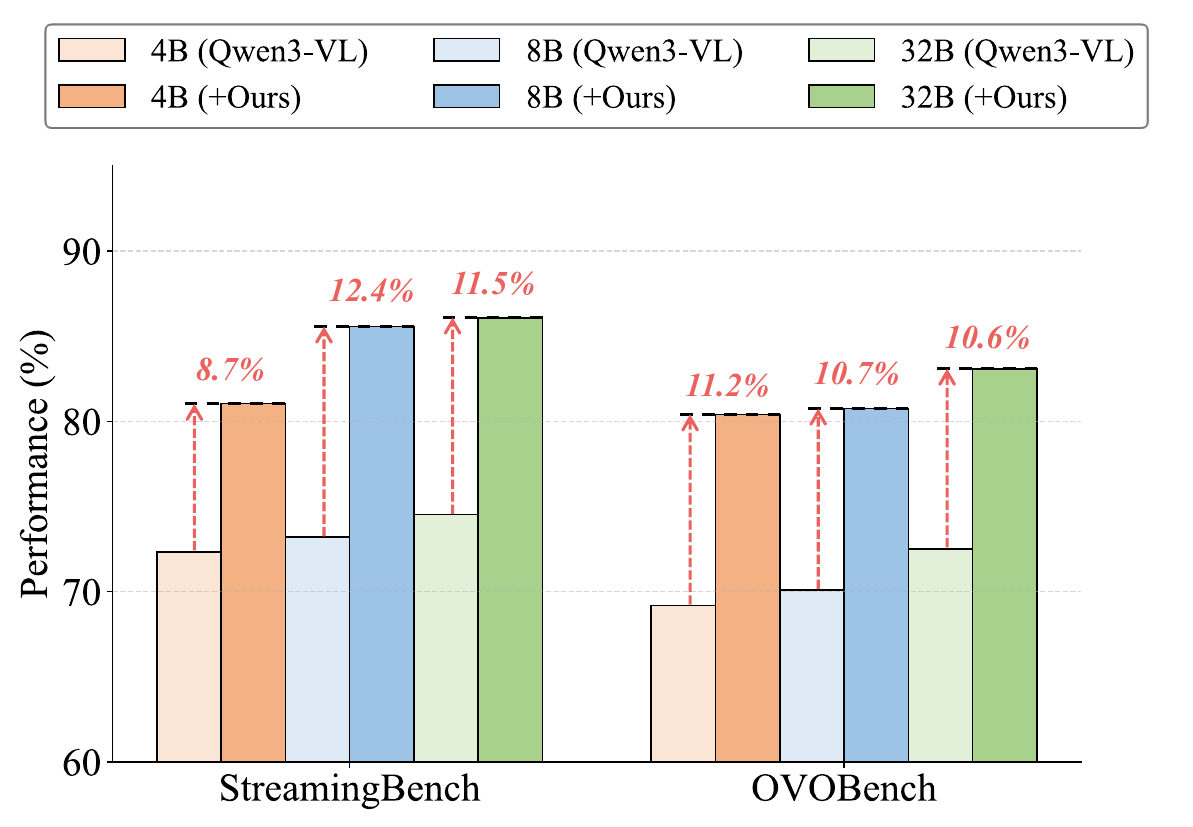}%
    \label{fig:qwen3_scaling}%
}
\hfil 
\subfloat[Impact of High-Res Ratio]{%
    \includegraphics[width=0.48\textwidth]{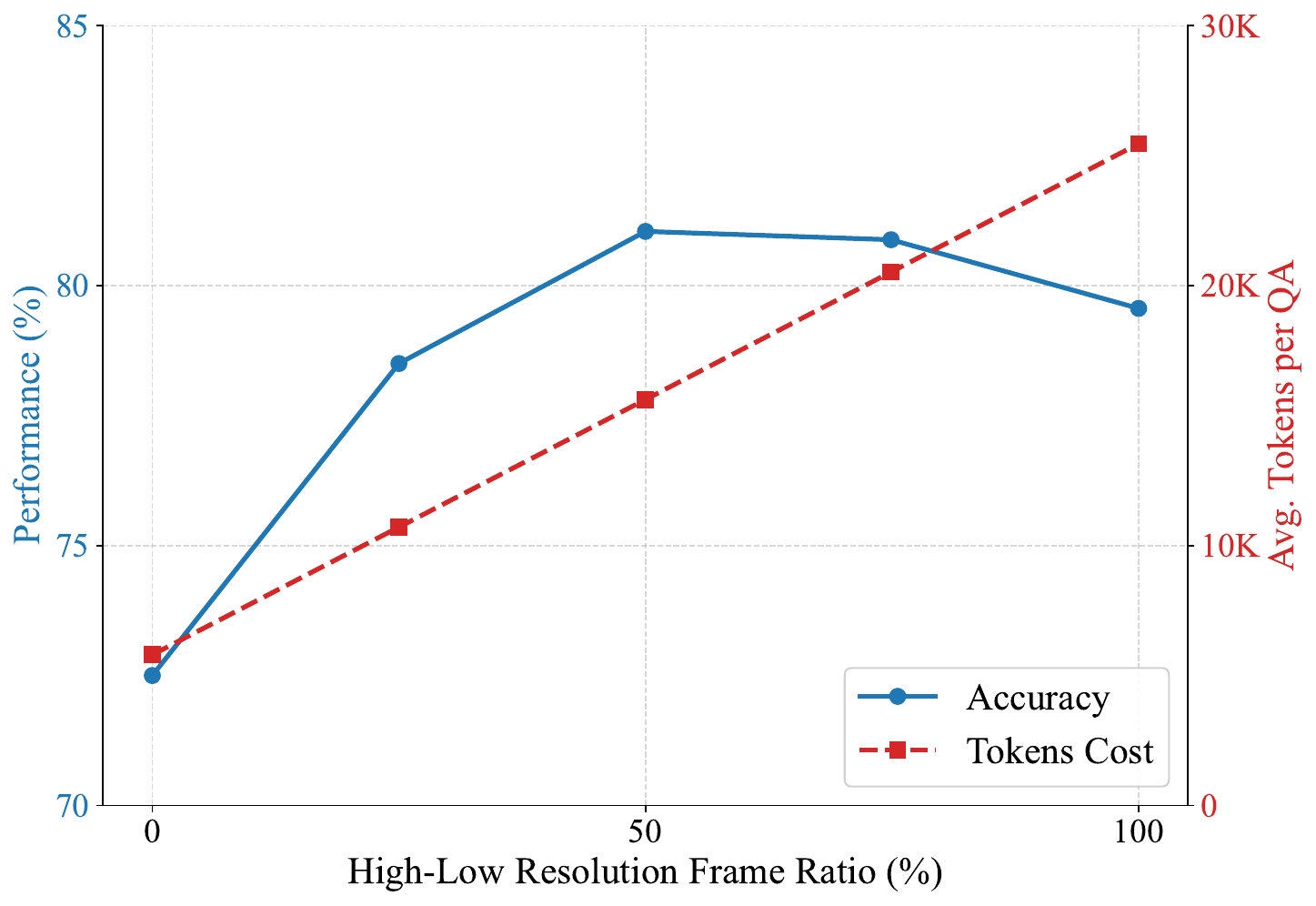}%
    \label{fig:impact}%
}
\caption{Scalability and memory allocation analysis. (a) CurveStream consistently delivers significant performance gains across varying model capacities (4B, 8B, 32B) of the Qwen3-VL series. (b) Impact of the clear memory (High-Res) retention ratio on overall accuracy and token cost. An adaptive $\sim$50\% ratio achieves the optimal trade-off between semantic integrity and computational overhead.}
\label{fig:overall_scalability} 
\end{figure*}

\cref{fig:qwen3_scaling} presents the evaluation results of our framework across the Qwen3-VL series with different parameter scales. Taking StreamingBench and OVOBench as examples, after being integrated into the 4B, 8B, and 32B versions of Qwen3-VL, our method yields absolute performance improvements of 8.7\%, 12.4\%, and 11.5\% on StreamingBench, respectively, compared to their corresponding uniform sampling baselines. Similarly, it achieves robust gains of 11.2\%, 10.7\%, and 10.6\% on OVOBench.

\begin{figure*}[tb]
    \centering
    \includegraphics[width=\textwidth]{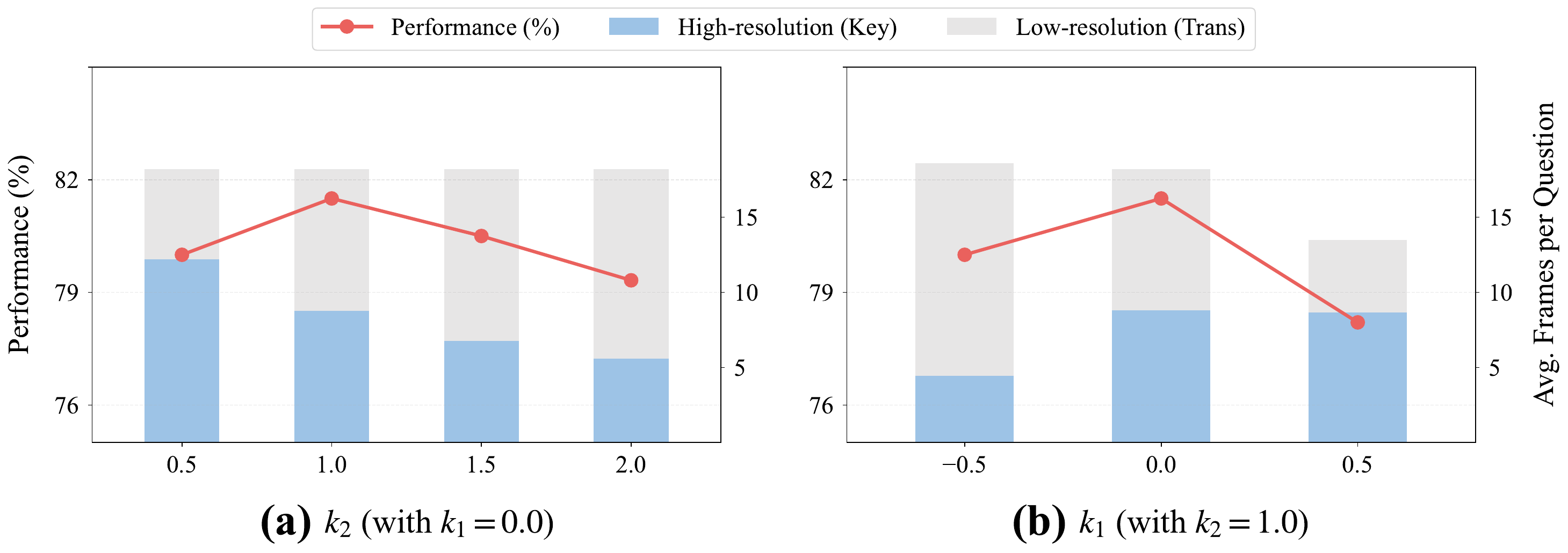}
    \caption{Ablation on K-Sigma dual thresholds. CurveStream exhibits strong hyperparameter robustness across various $k_1$ and $k_2$ configurations on OVOBench. The dynamic mechanism effectively balances memory allocation between High-Res and Low-Res frames, ensuring an optimal accuracy-efficiency trade-off without tedious tuning.}
    \label{ksigma_score}
\end{figure*}

These consistent quantitative improvements fully demonstrate that our curvature-aware adaptive memory mechanism does not overfit to models of a specific parameter volume. Instead, as a plug-and-play module, it maintains stable positive gains across multimodal base models ranging from small to large parameters, exhibiting exceptionally strong architectural universality and scalability.

\setcounter{table}{2} 
\begin{table*}[!t]
    \centering
    \begin{minipage}[t]{0.48\textwidth}
        \centering
        \caption{Comparison of frame sampling strategies using Qwen2-VL-7B under identical token constraints (N=10) on StreamingBench. Our metric achieves optimal performance by geometrically locating global semantic transitions.}
        \label{tab:table3curve}
        \vspace{0.2cm}
        \begin{tabular}{lc}
            \toprule
            \textbf{Sampling Strategy} & \textbf{Accuracy (\%)} \\
            \midrule
            Uniform Sampling & 69.04 \\
            Cosine Similarity & 73.28 \\
            Optical Flow & 46.54 \\
            Pyramid Optical Flow & 75.69 \\
            Streamforest (train) & 77.26 \\
            \midrule
            \textbf{Ours (Curvature)} & \textbf{77.31} \\
            \bottomrule
        \end{tabular}
    \end{minipage}\hfill
    \begin{minipage}[t]{0.48\textwidth}
        \centering
        \caption{Ablation on curvature score weights ($\lambda$) using Qwen2-VL-7B under identical token constraints on OVOBench. CurveStream maintains stable high performance across diverse weights, validating its plug-and-play reliability.}
        \label{tab:lamuda_score}
        \vspace{0.2cm}
        \begin{tabular}{lcc}
            \toprule
            \textbf{Method} & \textbf{$\lambda$} & \textbf{Accuracy (\%)} \\
            \midrule
            Qwen2-VL-7B & - & 60.65 \\
            \midrule
            Ours & 0.2 & \textbf{65.83} \\
            Ours & 0.4 & 62.50 \\
            Ours & 0.6 & 63.33 \\
            Ours & 0.8 & 62.50 \\
            Ours & 1.0 & 65.00 \\
            \bottomrule
        \end{tabular}
    \end{minipage}
\end{table*}

\subsection{Ablation Studies}
\label{sec:ablation}

To validate the independent contributions and synergistic effects of the core components in our adaptive memory framework, we conduct systematic ablation analyses on the Qwen-based model.

\textbf{Effectiveness of Curvature Metric.} To evaluate the superiority of manifold curvature in capturing temporal information increments, we compare different frame sampling strategies under identical visual token constraints (see Table \ref{tab:table3curve}). The results demonstrate that our curvature metric significantly outperforms both uniform sampling and motion sampling based on cosine similarity. This confirms that pure motion similarity struggles to distinguish redundant smooth-panning shots from sudden semantic shifts. Furthermore, compared to dense optical flow, which is computationally expensive and highly susceptible to pixel noise, temporal manifold curvature serves as a lightweight second-order geometric prior,enabling more precise and robust localization of core turning points.

\textbf{Adaptive Hierarchical Visual Memory Management.} We further ablate the allocation ratio between clear memory (native high-resolution keyframes) and blurred memory (down-projected low-resolution transition frames) in the memory queue. As illustrated in Fig. \ref{fig:impact}, forcing a 100\% clear memory strategy accelerates context window depletion, triggering catastrophic forgetting of early memory. Conversely, adopting a 0\% clear memory (``all-blur'') strategy discards critical spatial details, leading to a drastic performance drop. 

In contrast, the content-aware hybrid mechanism of our framework dynamically balances the clear memory ratio at approximately 50\% based on temporal dynamics. This approach achieves the best accuracy while substantially reducing computational overhead by about 40\%. This indicates that, compared to static constraints, dynamically allocating clear and blurred memory more effectively strikes a balance between the integrity of long-term context and the capture of fine-grained actions. Specifically, due to the temporal non-stationarity of streaming videos, forcing a fixed high-resolution retention often overfits to local translational motion noise. In contrast, our adaptive 50\% dynamic hybrid strategy essentially leverages localized blurred memory to serve as smooth transition states for action continuity, thereby freeing up the most critical clear memory space for high-curvature semantic transitions under the same token budget.

\textbf{Hyperparameter Robustness.} To verify the generalization stability of our framework, we evaluate the model's sensitivity to core hyperparameters. As shown in Table \ref{tab:lamuda_score}, when the curvature comprehensive score weight $\lambda$ varies across a broad range of $[0.1, 0.4]$, the model accuracy remains steadily above 62.5\%, peaking at 65.83\%. The maximum absolute fluctuation is merely 3.33\%, and it consistently outperforms the baseline method. Similarly, the dual-threshold parameters, $K\_SIGMA\_KEY$ ($k_1$) and $K\_SIGMA\_TRANS$ ($k_2$), maintain highly stable performance and robust frame sampling ratios across different settings (see Fig. \ref{ksigma_score}). Such exceptionally low hyperparameter sensitivity strongly corroborates the intrinsic robustness of our framework as a plug-and-play module, capable of adapting to diverse underlying data streams without tedious heuristic tuning for real-world streaming tasks.

%% file: sec/5-conclusion.tex
\section{Conclusion}
\label{sec:conclusion}

We present CurveStream, a training-free hierarchical memory management framework to boost streaming video understanding in MLLMs by tackling the inherent token explosion and Out-of-Memory (OOM) bottlenecks. Driven by the geometric insight that high-curvature regions in feature trajectories align with critical semantic transitions, CurveStream integrates a real-time Curvature Score with an online K-Sigma threshold. This dynamic mechanism adaptively routes incoming frames into clear or fuzzy memory states, ensuring MLLMs retain essential long-term visual context under strict token budgets. 

Extensive experiments demonstrate that this lightweight, model-agnostic module exhibits broad architectural compatibility and consistently yields substantial performance gains over respective baselines. By establishing new state-of-the-art results on challenging benchmarks like StreamingBench and OVOBench, CurveStream offers a robust solution for continuous video perception. Future work will extend this geometric memory paradigm to broader embodied AI applications, such as autonomous navigation and prolonged robotic manipulation, where real-time adaptive reasoning and decision-making are paramount.

%% file: sec/appendix.tex
\clearpage


\section{CurveStream Algorithm}
\label{sec:app_algorithm}

In this section, we provide the detailed pseudo-code for the proposed CurveStream framework. As outlined in \textbf{Algorithm~\ref{alg:curvestream}}, the online memory scheduling process operates sequentially on the incoming video stream without requiring any future context. For each new frame, the system first extracts its feature representation via the frozen visual encoder. Subsequently, the Curvature-Aware Scorer (CAS) evaluates the semantic transition by calculating the feature manifold curvature. Based on this dynamic curvature score and the recursively updated transient distribution, the Hierarchical Visual Memory Management (HVMM) module dynamically routes the current frame into either high-resolution \textit{Clear Memory} or down-sampled \textit{Blurred Memory} using dual adaptive thresholds. Finally, a strict First-In-First-Out (FIFO) eviction policy is applied to ensure the maximum memory footprint is strictly bounded.

\section{Qualitative Case Studies}
\label{sec:case_study}

\begin{figure*}[!t]
    \centering
    
    \includegraphics[width=0.8\textwidth]{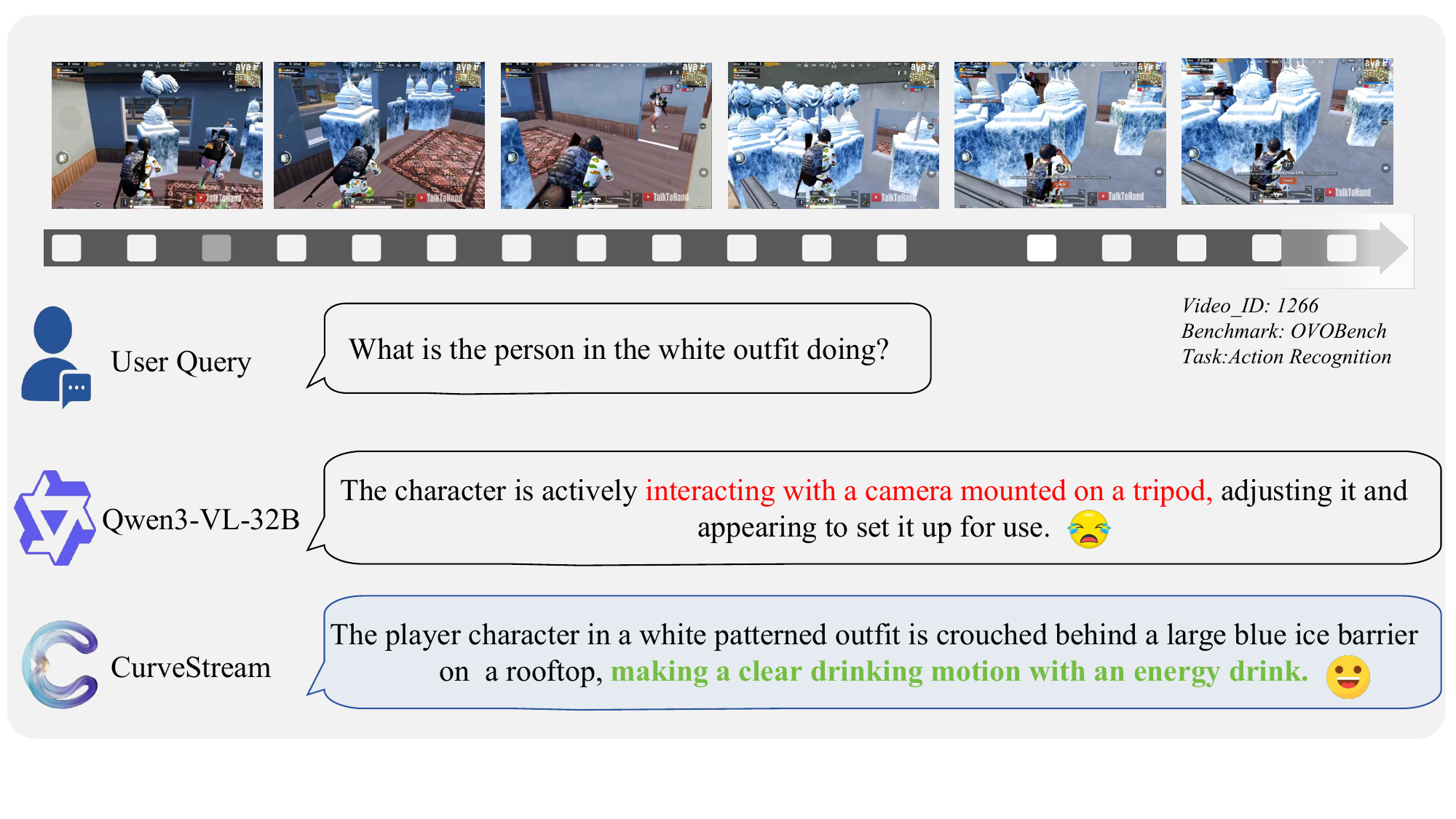}
    \caption{\textbf{Action Recognition in dynamic virtual environments.} Fast-paced viewpoint shifts often cause baseline models to lose track of transient actions, resulting in severe hallucinations (e.g., misinterpreting the action as setting up a camera). CurveStream captures the sharp curvature peak during the ``drinking'' animation, preserving it as a key semantic node to deliver an accurate response.}
    \label{fig:case_action}
    
    \vspace{0.5cm} 
    
    \includegraphics[width=0.8\textwidth]{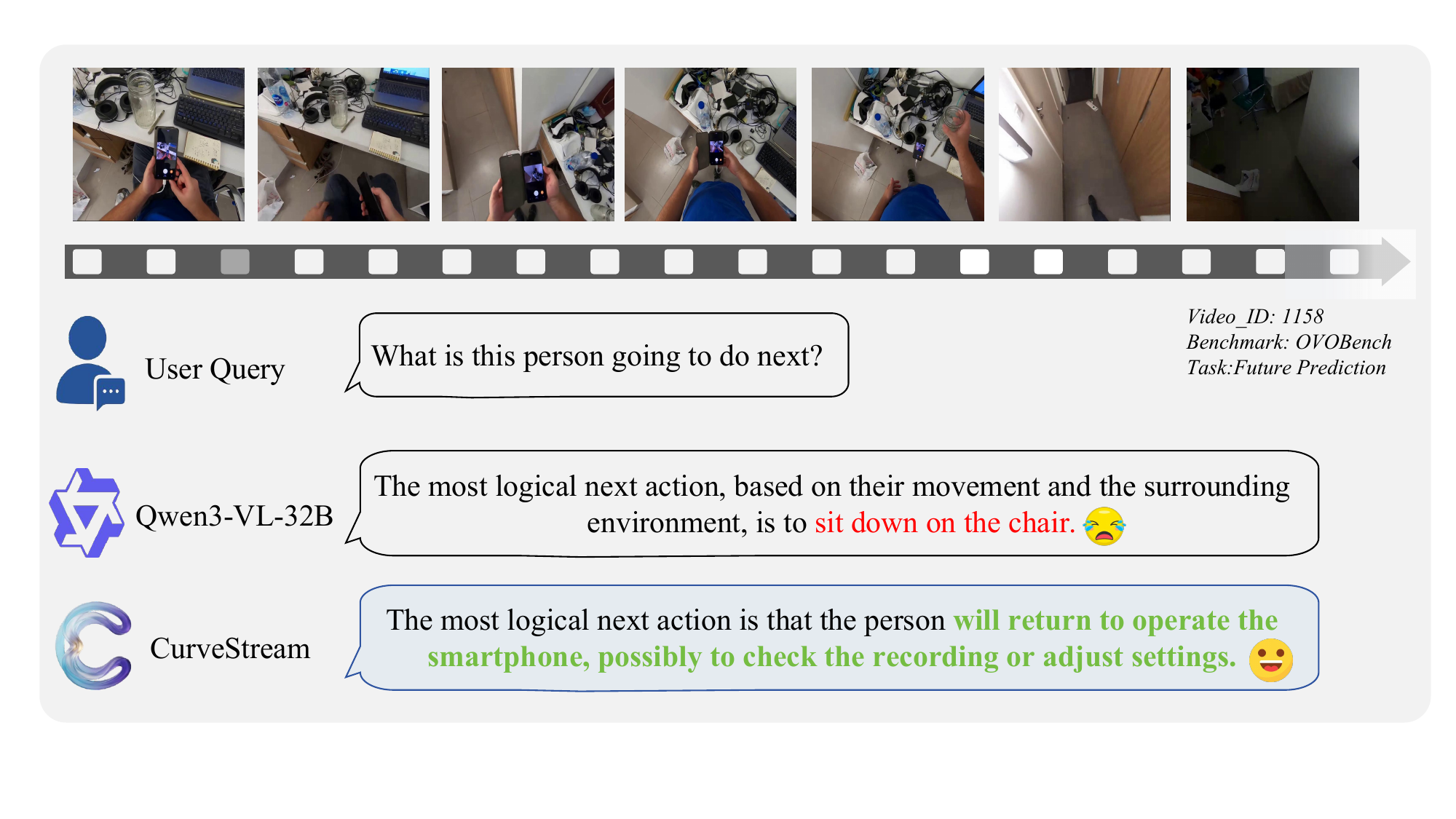}
    \caption{\textbf{Future Prediction in egocentric views.} Predicting future actions requires a complete and unbroken causal chain of past events. While the baseline suffers from context truncation and guesses the next action based on a biased background bias (the chair), CurveStream maintains a coherent sequence of the subject's interactions, correctly inferring the intention to operate the smartphone.}
    \label{fig:case_future}
    
\end{figure*}

\begin{figure*}[!t]
    \centering
    
    \includegraphics[width=0.8\textwidth]{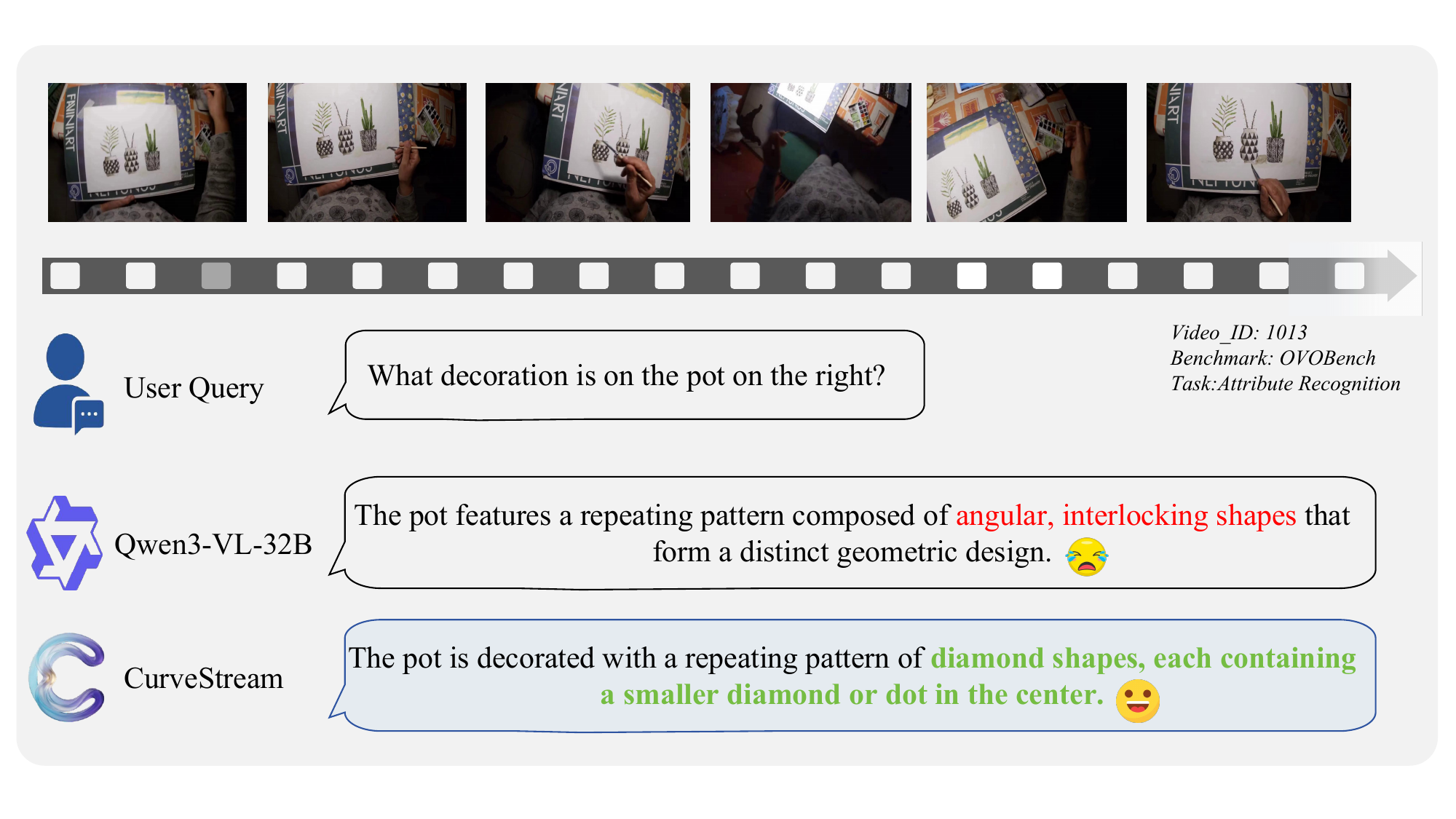}
    \caption{\textbf{Attribute Recognition requiring fine-grained spatial details.} Standard memory limits often force base models to downsample past frames uniformly, blurring complex textures. CurveStream dynamically assigns high-resolution Clear Memory to informative frames where the pot's pattern is unobscured, allowing it to correctly identify the nested diamond shapes.}
    \label{fig:case_attribute}
    
    \vspace{0.5cm} 
    
    \includegraphics[width=0.8\textwidth]{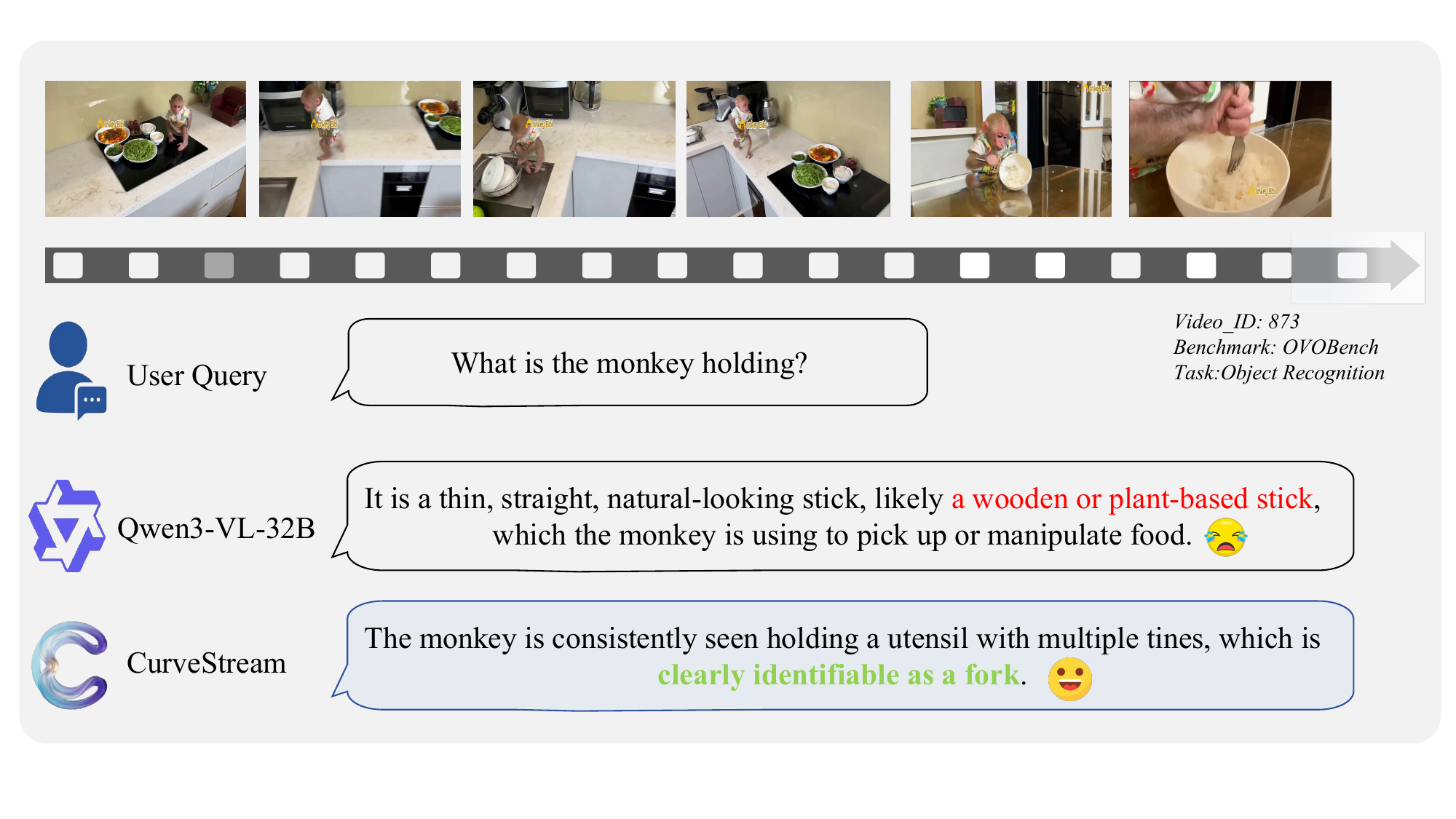}
    \caption{\textbf{Object Recognition under severe occlusion.} Tracking small objects (like the fork) is notoriously difficult in long videos. CurveStream registers the semantic shift when the utensil is clearly exposed, safeguarding this vital visual evidence in the memory queue to prevent the baseline's ``wooden stick'' hallucination.}
    \label{fig:case_object}

\end{figure*}

To intuitively illustrate the effectiveness of our memory mechanism in handling complex, unconstrained streaming videos, we provide qualitative comparisons between CurveStream and the robust baseline model (Qwen3-VL-32B) in Fig.~\ref{fig:case_action} to Fig.~\ref{fig:case_object}. We select four highly challenging sub-tasks from OVOBench: Action Recognition (Fig.~\ref{fig:case_action}), Future Prediction (Fig.~\ref{fig:case_future}), Attribute Recognition (Fig.~\ref{fig:case_attribute}), and Object Recognition (Fig.~\ref{fig:case_object}). 

In highly dynamic or visually cluttered scenarios, standard MLLMs often suffer from severe hallucination or catastrophic forgetting. This is primarily because their passive memory eviction policies indiscriminately discard historical tokens, leading to broken causal chains, or their uniform downsampling strategies irreparably blur fine-grained spatial details. As demonstrated in the following cases, CurveStream successfully overcomes these bottlenecks. By monitoring the feature manifold curvature, our framework accurately anchors critical semantic transitions (e.g., the sudden appearance of a small object or a rapid action shift) and intelligently routes them into the high-resolution Clear Memory. This ensures the model maintains a precise, coherent, and hallucination-free understanding across the entire streaming timeline.

\begin{algorithm}[tb]
\caption{CurveStream: Curvature-Aware Hierarchical Visual Memory Management}
\label{alg:curvestream}
\begin{algorithmic}[1]
\REQUIRE Continuous video stream $\mathcal{V}=\{I_t\}_{t=1}^\infty$ and query timestamp $t_q$; target memory capacity $N_{max}$; balancing coefficient $\lambda$; threshold multipliers $k_1, k_2$ $(k_1 < k_2)$
\ENSURE An adaptively updated visual memory queue $\mathcal{M}_t$

\STATE $\triangleright$ Initialize the memory queue $\mathcal{M}_0 \leftarrow \emptyset$, and the time step $t \leftarrow 1$
\STATE Initialize transient distribution parameters: $\mu_0 \leftarrow 0$, $\sigma_0 \leftarrow 0$
\WHILE{receiving incoming frame $I_t$ from stream $\mathcal{V}$}
    \STATE $\triangleright$ Extract and $L_2$-normalize global feature representation $F_t \in \mathbb{R}^D$ via the frozen visual encoder
    \IF{$t \ge 3$}
        \STATE $\triangleright$ Stage 1: Curvature-Aware Scorer (CAS)
        \STATE Compute first-order Motion Variation: $M_t = \mathcal{D}_{motion}(F_t, F_{t-1})$
        \STATE Compute second-order Geometric Curvature: $C_t = \mathcal{K}_{geo}(F_t, F_{t-1}, F_{t-2})$
        \STATE Calculate the final Curvature Score: $CS_t = M_t + \lambda C_t$
        
        \vspace{3pt}
        \STATE $\triangleright$ Stage 2: Hierarchical Visual Memory Management (HVMM)
        \STATE Recursively update the transient manifold distribution state:
        \STATE $(\mu_t, \sigma_t) \leftarrow \text{UpdateDistributionState}(CS_t, \mu_{t-1}, \sigma_{t-1})$
        \STATE Generate dynamic dual thresholds: 
        \STATE $g_1, g_2 \leftarrow \text{CalculateDynamicThresholds}(\mu_t, \sigma_t, k_1, k_2)$
        
        \vspace{3pt}
        \IF{$CS_t \ge g_2 \textbf{ or } t == t_q$}
            \STATE $\triangleright$ Retain as Clear Memory to capture significant semantic shifts
            \STATE $\mathcal{M}_t = \text{Update}(\mathcal{M}_{t-1}, I_t, s_t=\text{Clear}, r_t=\text{High})$
        \ELSIF{$g_1 \le CS_t < g_2$}
            \STATE $\triangleright$ Retain as Blurred Memory for intermediate transition states
            \STATE $\mathcal{M}_t = \text{Update}(\mathcal{M}_{t-1}, I_t, s_t=\text{Blurred}, r_t=\text{Low})$
        \ELSE
            \STATE $\triangleright$ Discard low-information redundant features
            \STATE $\mathcal{M}_t = \mathcal{M}_{t-1}$
        \ENDIF
        
        \vspace{3pt}
        \IF{$|\mathcal{M}_t| > N_{max}$}
            \STATE $\triangleright$ Execute strict First-In-First-Out (FIFO) eviction
            \STATE Remove the oldest tokens from $\mathcal{M}_t$
        \ENDIF
    \ENDIF
    \STATE $\triangleright$ $t \leftarrow t + 1$
\ENDWHILE
\end{algorithmic}
\end{algorithm}

\section{Theoretical Analysis of the Geometric Curvature Metric}
\label{sec:app_curvature_analysis}

In this section, we provide a rigorous theoretical formulation for the geometric curvature ($C_t$) metric introduced in the Curvature-Aware Scorer (CAS). From a discrete geometric perspective, we demonstrate how this metric theoretically decouples core semantic transitions from continuous physical motion noise.

\subsection{Kinematic Modeling in the Latent Manifold}

Let the continuous video stream be mapped into a high-dimensional latent feature space. Following $L_2$ normalization, the observation of each video frame $I_t$ is projected onto a unit hypersphere, yielding the feature representation $F_t \in \mathbb{R}^D$. The temporal evolution of the video stream constructs a discrete parameterized curve on this hyperspherical manifold.

From a kinematic perspective, the first-order feature displacement vectors $d_1 = F_{t-1} - F_{t-2}$ and $d_2 = F_t - F_{t-1}$ represent the discrete velocity vectors of the visual signal at adjacent time steps. Traditional similarity metrics (\eg, inter-frame cosine similarity) primarily rely on the magnitude of these velocity vectors, which inherently conflates semantic transitions with smooth, continuous camera motions (\eg, panning).

\subsection{Differential Geometric Perspective of $C_t$}

To isolate semantic intensity, we approximate the second-order geometric curvature of the feature trajectory. In continuous differential geometry, the curvature $\kappa$ of a parameterized curve measures the rate of change of the unit tangent vector with respect to arc length.

We map this definition onto our discrete manifold. First, we compute the unit tangent vectors (\ie, normalized velocity vectors) at adjacent time steps:
\begin{equation}
    T_1 = \frac{d_1}{||d_1||}, \quad T_2 = \frac{d_2}{||d_2||}
\end{equation}

The geometric curvature metric proposed in this paper is defined as the cosine distance between adjacent displacement vectors:
\begin{equation}
    C_t = 1 - \frac{\langle d_1, d_2 \rangle}{||d_1|| \cdot ||d_2||} = 1 - \langle T_1, T_2 \rangle
\end{equation}

In Euclidean space, the squared distance between two unit vectors has a strict mathematical equivalence with their inner product:
\begin{equation}
    ||T_2 - T_1||^2 = ||T_2||^2 + ||T_1||^2 - 2\langle T_1, T_2 \rangle = 2(1 - \langle T_1, T_2 \rangle)
\end{equation}

Substituting this into our metric yields the geometric equivalence:

\begin{equation}
    C_t = \frac{1}{2} ||T_2 - T_1||^2
\end{equation}

This theoretical derivation proves that $C_t$ is strictly equivalent (up to a constant scaling factor) to the squared variation of the unit tangent vector.Thus, as a discrete approximation of manifold curvature, $C_t$ geometrically evaluates the directional derivative of feature evolution instead of a mere scalar displacement.

\subsection{Theoretical Advantages of Semantic Decoupling}

This curvature-based formulation inherently provides two critical theoretical advantages for streaming video understanding:

\begin{itemize}
    \item[$\bullet$] \textbf{Immunity to Constant Velocity Motion Noise:} In scenarios with smooth, continuous motion (\eg, stable camera panning), the feature trajectory evolves at a relatively constant velocity. Geometrically, its tangent vectors remain approximately parallel ($T_1 \approx T_2$), yielding $\langle T_1, T_2 \rangle \approx 1$ and $C_t \approx 0$. Consequently, this geometric penalty inherently suppresses low-level physical motion noise by mechanism.
    
    \item[$\bullet$] \textbf{Orthogonal Sensitivity to Semantic Transitions:} When sudden semantic shifts occur (\eg, shot changes, new entities entering the frame, or sharp action boundaries), the feature trajectory undergoes a drastic directional deviation. The new velocity vector $d_2$ is projected into a subspace that is nearly orthogonal or even divergent from $d_1$. This forces the inner product $\langle T_1, T_2 \rangle$ to drop sharply, thereby generating a distinct curvature spike.
\end{itemize}

By introducing this second-order geometric prior, the CAS module achieves an effective decoupling of core semantic transitions from redundant background dynamics on a mathematical basis, laying a robust theoretical foundation for the subsequent K-Sigma dynamic memory routing mechanism.
\section{Detailed Performance  on Streaming Benchmarks}
\label{sec:app_online_perf}

We present the comprehensive, fine-grained evaluation results of our proposed curvature-aware hierarchical visual memory management method on streaming video benchmarks. The detailed breakdowns across StreamingBench~\cite{lin2024streamingbench} (\textbf{Table~\ref{tab:app_streamingbench_detailed}}) and OVO-Bench~\cite{niu2025ovo} (\textbf{Table~\ref{tab:app_ovobench_detailed}}) are shown below. We compare our approach against standard MLLMs (\eg, Qwen2.5-VL~\cite{Qwen2.5-VL}) and state-of-the-art streaming baselines (\eg, StreamForest~\cite{zeng2025streamforest}).
\begin{table*}[!ht]
  \centering
  \caption{Comprehensive evaluation results on the \textbf{StreamingBench} benchmark for real-time visual understanding. Our method consistently improves the performance of base MLLMs across all contextual reasoning and event understanding metrics, as indicated by the ``$\uparrow$'' symbol. The highest scores in each column are marked in \textbf{bold}.}
  \label{tab:app_streamingbench_detailed}
  \resizebox{\textwidth}{!}{
  \begin{tabular}{lcccccccccccc}
    \toprule
    Model & Frame & OP & CR & CS & ATP & EU & TR & PR & SU & ACP & CT & Avg. \\
    \midrule
    Human & - & 89.47 & 92.00 & 93.60 & 91.47 & 95.65 & 92.52 & 88.00 & 88.75 & 89.74 & 91.30 & 91.46 \\
    \midrule
    \rowcolor{gray!10}\multicolumn{13}{l}{\textit{Proprietary MLLMs}} \\
    \midrule
    Gemini 1.5 Pro~\cite{gemini} & 1 fps & 79.02 & 80.47 & 83.54 & 79.67 & 80.00 & 84.74 & 77.78 & 64.23 & 71.95 & 48.70 & 75.69 \\
    GPT-4o~\cite{hurst2024gpt4o} & 64 & 77.11 & 80.47 & 83.91 & 76.47 & 70.19 & 83.80 & 66.67 & 62.19 & 69.12 & 49.22 & 73.28 \\
    Claude 3.5 Sonnet~\cite{claude35sonnet} & 20 & 73.33 & 80.47 & 84.09 & 82.02 & 75.39 & 79.53 & 61.11 & 61.79 & 69.32 & 43.09 & 72.44 \\
    \midrule
    \rowcolor{gray!10}\multicolumn{13}{l}{\textit{Open-source Offline MLLMs}} \\
    \midrule
    Qwen2-VL-7B~\cite{wang2024qwen2} & 32 & 55.86 & 55.47 & 57.41 & 58.17 & 52.80 & 43.61 & 39.81 & 42.68 & 45.61 & 35.23 & 49.52 \\
    InternVL-V2-8B~\cite{chen2024internvl} & 14 & 53.68 & 49.22 & 70.98 & 56.86 & 53.42 & 53.89 & 54.63 & 48.78 & 50.14 & 17.62 & 52.32 \\
    \midrule
    \rowcolor{gray!10}\multicolumn{13}{l}{\textit{Open-source Online MLLMs}} \\
    \midrule
    Flash-VStream-7B~\cite{zhang2025flash} & - & 25.89 & 43.57 & 24.91 & 23.87 & 27.33 & 13.08 & 18.52 & 25.20 & 23.87 & 48.70 & 23.23 \\
    VideoLLM-online-8B~\cite{chen2024videollmonline} & 2 fps & 39.07 & 40.06 & 34.49 & 31.05 & 45.96 & 32.40 & 31.48 & 34.16 & 42.49 & 27.89 & 35.99 \\
    Dispider-7B~\cite{qian2025dispider} & 1 fps & 74.92 & 75.53 & 74.10 & 73.08 & 74.44 & 59.92 & 76.14 & 62.91 & 62.16 & 45.80 & 67.63 \\
    TimeChat-Online-7B~\cite{yao2025timechat} & 1 fps & 80.22 & 82.03 & 79.50 & 83.33 & 76.10 & 78.50 & 78.70 & 64.63 & 69.60 & 57.98 & 75.36 \\
    StreamForest-7B~\cite{zeng2025streamforest} & 1 fps & 83.11 & 82.81 & 82.65 & 84.26 & 77.50 & 78.19 & 76.85 & 69.11 & 75.64 & 54.40 & 77.26 \\
    \midrule
    \rowcolor{gray!10}\multicolumn{13}{l}{\textit{Training-free Offline-to-Online Methods}} \\
    \midrule
    LLaVA-OV-7B~\cite{li2024llava} & 32 & 78.75 & 78.12 & 80.76 & 81.19 & 71.70 & 72.59 & 72.22 & 63.82 & 66.01 & 38.34 & 71.34 \\
    \quad + ReKV~\cite{di2025rekv}  & 0.5 fps & 76.02 & 81.25 & 77.92 & 76.90 & 66.04 & 66.04 & 69.44 & 60.98 & 64.31 & 49.22 & 69.22 \\
    \quad + HERMES~\cite{zhang2026hermeskvcachehierarchical}  & 0.5 fps & 79.02 & 81.25 & 87.70 & 80.20 & 69.18 & 71.96 & 73.15 & 66.26 & 69.41 & 43.52 & 73.23 \\
    \rowcolor{blue!5} \quad \textbf{+ Ours (CurveStream)} & 10-20 & \textbf{85.56} & \textbf{85.13} & 71.88 & \textbf{88.52} & \textbf{72.50} & \textbf{83.49} & 65.74 & \textbf{69.51} & \textbf{67.90} & 35.42 & \textbf{75.12} (\textcolor{red}{$\uparrow$ \textbf{3.78}}) \\
    \midrule
    Qwen2-VL-7B~\cite{wang2024qwen2} & 1 fps & 77.38 & 76.56 & 73.19 & 75.08 & 75.00 & 67.91 & 73.15 & 65.04 & 66.57 & 35.75 & 69.04 \\
    \quad + Freshmem~\cite{li2026freshmem} & 1 fps & 84.47 & 83.59 & 77.60 & 83.28 & 78.12 & 80.37 & 70.37 & 74.39 & 66.86 & 30.05 & 74.20 \\
    \rowcolor{blue!5} \quad \textbf{+ Ours (CurveStream)} & 10-20 & \textbf{88.56} & \textbf{77.34} & \textbf{88.61} & \textbf{89.84} & \textbf{76.25} & \textbf{92.52} & \textbf{76.85} & \textbf{76.83} & \textbf{76.70} & \textbf{45.31} & \textbf{81.04} (\textcolor{red}{$\uparrow$ \textbf{12.00}}) \\
    \midrule
    Qwen2.5-VL-7B~\cite{Qwen2.5-VL} & 1 fps & 77.93 & 76.56 & 78.55 & 80.86 & 76.73 & 76.95 & 80.56 & 65.45 & 65.72 & 52.85 & 73.31 \\
    \quad + HERMES~\cite{zhang2026hermeskvcachehierarchical} & 1 fps & 83.65 & 81.25 & 88.01 & 87.46 & 76.73 & 86.60 & 82.41 & 76.02 & 73.94 & 46.63 & 79.44 \\
    \rowcolor{blue!5} \quad \textbf{+ Ours (CurveStream)} & 10-20 & \textbf{90.19} & \textbf{78.12} & \textbf{94.94} & \textbf{89.51} & \textbf{81.25} & \textbf{95.02} & \textbf{83.33} & \textbf{83.74} & \textbf{79.26} & 44.79 & \textbf{84.00} (\textcolor{red}{$\uparrow$ \textbf{10.69}}) \\
    \midrule
    Qwen3-VL-8B~\cite{bai2025qwen3} & 1 fps & 76.84 & 77.22 & 77.29 & 80.74 & 70.35 & 75.21 & 80.56 & 64.23 & 65.76 & 49.22 & 73.2 \\
    \rowcolor{blue!5} \quad \textbf{+ Ours (CurveStream)} & 10-20 & \textbf{90.74} & \textbf{79.69} & \textbf{95.25} & \textbf{93.44} & \textbf{81.88} & \textbf{95.95} & \textbf{85.19} & \textbf{79.27} & \textbf{85.23} & 47.92 & \textbf{85.56} (\textcolor{red}{$\uparrow$ \textbf{12.36}}) \\
    \bottomrule
  \end{tabular}
  }
\end{table*}

\begin{table*}[!ht]
  \centering
  \caption{Detailed performance comparison on the \textbf{OVOBench} dataset across various real-time visual perception sub-tasks. We report the evaluation metric (\eg, Accuracy \%) for both the standard base models and our proposed CurveStream. The ``$\uparrow$'' denotes the absolute performance gain achieved by integrating our curvature-aware memory management into the respective base models. Best results are highlighted in \textbf{bold}.}
  \label{tab:app_ovobench_detailed}
  \resizebox{\textwidth}{!}{
  \begin{tabular}{lcccccccc}
    \toprule
    Model & Frame & OCR & ACR & ATR & STU & FPD & OJR & Avg. \\
    \midrule
    Human & - & 93.96 & 92.57 & 94.83 & 92.70 & 91.09 & 94.02 & 93.20 \\
    \midrule
    \rowcolor{gray!10}\multicolumn{9}{l}{\textit{Proprietary MLLMs}} \\
    \midrule
    Gemini 1.5 Pro~\cite{gemini} & 1 fps & 85.91 & 66.97 & 79.31 & 58.43 & 63.37 & 61.96 & 69.32 \\
    GPT-4o~\cite{hurst2024gpt4o} & 64 & 69.80 & 64.22 & 71.55 & 51.12 & 70.30 & 59.78 & 64.46 \\
    \midrule
    \rowcolor{gray!10}\multicolumn{9}{l}{\textit{Open-source Offline MLLMs}} \\
    \midrule
    LLaVA-Video-7B~\cite{li2024llava} & 64 & 69.80 & 59.63 & 66.38 & 50.56 & 72.28 & 61.41 & 63.34 \\
    Qwen2-VL-7B~\cite{wang2024qwen2} & 64 & 69.13 & 53.21 & 63.79 & 50.56 & 66.34 & 60.87 & 60.65 \\
    InternVL2-8B~\cite{chen2024internvl} & 64 & 68.46 & 58.72 & 68.97 & 44.94 & 67.33 & 55.98 & 60.73 \\
    LongVU-7B & 1 fps & 55.70 & 49.54 & 59.48 & 48.31 & 68.32 & 63.04 & 57.40 \\
    \midrule
    \rowcolor{gray!10}\multicolumn{9}{l}{\textit{Open-source Online MLLMs}} \\
    \midrule
    VideoLLM-online-8B~\cite{chen2024videollmonline} & 2 fps & 8.05 & 23.85 & 12.07 & 14.04 & 45.54 & 21.20 & 20.79 \\
    Flash-VStream-7B~\cite{zhang2025flash} & 1 fps & 25.50 & 32.11 & 29.31 & 33.71 & 29.70 & 28.80 & 29.86 \\
    Dispider-7B~\cite{qian2025dispider} & 1 fps & 57.72 & 49.54 & 62.07 & 44.94 & 61.39 & 51.63 & 54.55 \\
    TimeChat-Online-7B~\cite{yao2025timechat} & 1 fps & 75.20 & 46.80 & 70.70 & 47.80 & 69.30 & 61.40 & 61.90 \\
    StreamForest-7B~\cite{zeng2025streamforest} & 1 fps & 68.46 & 53.21 & 71.55 & 47.75 & 65.35 & 60.87 & 61.20 \\
    \midrule
    \rowcolor{gray!10}\multicolumn{9}{l}{\textit{Training-free Offline-to-Online Methods}} \\
    \midrule
    LLaVA-OV-7B~\cite{li2024llava} & 32 & 67.79 & 55.05 & 72.41 & 48.31 & 72.28 & 62.50 & 63.06 \\
    \quad + ReKV~\cite{di2025rekv} & 0.5 fps & 52.35 & 54.13 & 69.83 & 43.26 & 67.33 & 57.07 & 57.33 \\
    \quad + HERMES~\cite{zhang2026hermeskvcachehierarchical} & 0.5 fps & 72.48 & 62.39 & 74.14 & 50.56 & 73.27 & 65.22 & 66.34 \\
    \rowcolor{blue!5} \quad \textbf{+ Ours (CurveStream)} & 10-20 & \textbf{84.56} & \textbf{66.97} & \textbf{77.59} & \textbf{53.93} & \textbf{74.26} & \textbf{70.65} & \textbf{70.57} (\textcolor{red}{$\uparrow$ \textbf{7.51}}) \\
    \midrule
    Qwen2-VL-7B~\cite{wang2024qwen2} & 1 fps & 69.13 & 53.21 & 63.79 & 50.56 & 66.34 & 60.87 & 60.65 \\
    \quad + Freshmem~\cite{li2026freshmem} & 1 fps & 77.18 & 60.55 & 70.69 & 56.74 & 63.37 & 70.65 & 66.67 \\
    \rowcolor{blue!5} \quad \textbf{+ Ours (CurveStream)} & 10-20 & \textbf{86.58} & \textbf{73.29} & \textbf{79.31} & 48.31 & \textbf{70.30} & \textbf{72.83} & \textbf{70.73} (\textcolor{red}{$\uparrow$ \textbf{10.08}}) \\
    \midrule
    Qwen2.5-VL-7B~\cite{Qwen2.5-VL} & 1 fps & 67.79 & 55.05 & 67.24 & 42.13 & 66.34 & 60.87 & 59.90 \\
    \quad + HERMES~\cite{zhang2026hermeskvcachehierarchical} & 0.5 fps & 85.23 & 64.22 & 71.55 & 53.37 & 74.26 & 65.22 & 68.98 \\
    \rowcolor{blue!5} \quad \textbf{+ Ours (CurveStream)} & 10-20 & \textbf{87.25} & \textbf{70.64} & \textbf{79.31} & \textbf{57.87} & \textbf{76.24} & \textbf{73.91} & \textbf{73.48} (\textcolor{red}{$\uparrow$ \textbf{13.58}}) \\
    \midrule
    Qwen3-VL-8B~\cite{bai2025qwen3} & 1 fps & 71.14 & 65.14 & 75.86 & 64.61 & 75.25 & 70.65 & 70.10 \\
    \rowcolor{blue!5} \quad \textbf{+ Ours (CurveStream)} & 10-20 & \textbf{93.96} & \textbf{82.57} & \textbf{83.62} & \textbf{68.54} & \textbf{78.22} & \textbf{80.43} & \textbf{80.76} (\textcolor{red}{$\uparrow$ \textbf{10.66}}) \\
    \bottomrule
  \end{tabular}
  }
\end{table*}


The comprehensive performance improvements of CurveStream across both streaming benchmarks are primarily attributed to our redesign of the Hierarchical Visual Memory Management mechanism. Confronted with the continuous growth of tokens in long streaming videos, base models are typically bounded by rigid memory mechanisms (\eg, fixed uniform downsampling or passive FIFO cache eviction). This easily leads to the loss of high-value semantic information and the disruption of the model's contextual coherence. CurveStream constructs an adaptive Hierarchical Visual Memory system. We utilize the local curvature on the feature manifold as a perceptual heuristic to guide the dynamic allocation of memory: under the premise of strictly constraining memory overhead, the video stream is intelligently decoupled into high-resolution Clear Memory and high-compression-ratio Blurred Memory. This strategy of ``semantic-perception-driven memory routing'' effectively alleviates the resource allocation bottlenecks of base models in long sequences, providing solid architectural support for the performance leaps across various sub-tasks.

\subsection{Analysis of Improvements on StreamingBench}
\label{subsec:analysis_streamingbench}

\textbf{CR (Causal Reasoning), EU (Event Understanding) \& ACP (Action Perception):} One of the core challenges of StreamingBench lies in memory retention under long-term contexts. Constrained by limited context windows, base models often have early key events squeezed out by subsequent redundant frames, leading to difficulties in long-range reasoning. CurveStream's hierarchical architecture provides a viable path to alleviate this issue. Clear Memory focuses on the persistent storage of discrete salient events triggered by high curvature, while Blurred Memory maintains the background context between events at a lower token cost. This macroscopic memory scheduling approach constructs a relatively complete and compact ``causal topological chain'' for the model, assisting it in better handling complex, long-range logical correlation problems even when operating under severely limited memory capacities.

\textbf{CT (Counting) \& CS (Clips Summarization):} In counting and summarization tasks, the loss of historical states is often a critical cause of model output errors. CurveStream's memory management demonstrates strong robustness here. By transforming significant action mutations into discrete keyframe snapshots and retaining them, it essentially compresses the continuous, lengthy video stream into a high-density sequence containing core events. This mechanism provides base models with a more structured and reliable basis for memory retrieval when handling complex frequency statistics, event counting, and global video summarization queries.

\subsection{Analysis of Improvements on OVO-Bench}
\label{subsec:analysis_ovobench}

\textbf{OCR (Optical Character Recognition) \& ATR (Attribute Recognition):} These tasks highly rely on the retention of high-resolution visual features. Under memory pressure, base models often resort to global downsampling, easily causing an irreversible loss of fine-grained information. CurveStream's hierarchical memory management adeptly tackles this resource allocation dilemma. When the Curvature-Aware Scorer (CAS) detects significant changes in text or attributes, the system prioritizes allocating the token budget to these key frames, maintaining their native high resolution as Clear Memory. Simultaneously, low-information-density background frames are compressed into Blurred Memory. This dynamic memory scheduling strategy significantly enhances the model's perception of fine-grained information while maintaining a highly stable and consistent overall memory footprint.

\textbf{ACR (Action Recognition) \& FPD (Future Prediction):} The sliding window memory mechanism of base models, when constrained by capacity, easily evicts the preceding states of actions, thereby compromising the integrity of temporal logic. CurveStream maps the fluctuations of actions to curvature variations on the feature manifold, utilizing these variations to assist in locating action boundaries and anchoring them as key semantic nodes in the working memory. This mechanism helps ensure that the model is supported by a more coherent and complete history of state transitions when reasoning about current actions or predicting future evolutions, effectively reducing the risk of hallucination caused by context truncation.

\textbf{STU (Spatial Understanding) \& OJR (Object Recognition):} Complex spatial structures and target poses constantly change with camera motion. Fixed uniform sampling strategies sometimes fail to retain frames with optimal viewpoints in memory. With the help of the K-Sigma dynamic threshold, CurveStream achieves adaptive memory updating, enabling the system to better adapt to variable camera motion rhythms. It maximizes the retention of frames containing rich spatial topological relations in the core memory area, thereby substantially reducing visual information omissions typically caused by improper or rigid memory scheduling.

\section{Generalization on Offline Video Understanding}
\label{sec:app_offline_perf}

Although the CurveStream architecture was primarily designed to alleviate memory bottlenecks in streaming scenarios, its core mechanism---Curvature-Aware Hierarchical Visual Memory Management also provides an efficient representation paradigm for offline long-video understanding. In the offline evaluation setting, confronted with complete video sequences, CurveStream overcomes the limitations of conventional fixed frame sampling. By evaluating the semantic information density across the global temporal axis and utilizing curvature to adaptively route the limited token budget to highly dynamic segments, this mechanism demonstrates highly robust generalization capabilities when evaluated across two major offline video understanding benchmarks. 

The details across FAVOR-Bench~\cite{tu2025favor} (\textbf{Table~\ref{tab:app_favorbench_detailed}}) and MVBench
~\cite{li2024mvbench} (\textbf{Table~\ref{tab:app_mvbench_detailed}}) are presented below.
\begin{table*}[!ht]
  \centering
  \caption{Detailed performance comparison on the \textbf{FavorBench} dataset. "$\uparrow$" indicates the performance improvement of our method compared to the base model.}
  \label{tab:app_favorbench_detailed}
  \resizebox{\textwidth}{!}{
  \begin{tabular}{lcccccccc}
    \toprule
    Model & Frame & AS & HAC & SAD & MAD & CM & NSM & Avg. \\
    \midrule
    \rowcolor{gray!10}\multicolumn{9}{l}{\textit{Proprietary MLLMs}} \\
    \midrule
    Gemini-1.5-Pro~\cite{gemini} & 1 fps* & 49.22 & 53.73 & 48.80 & 54.85 & 41.58 & 56.25 & 49.87 \\
    GPT-4o~\cite{hurst2024gpt4o} & 1 fps* & 40.65 & 45.10 & 42.84 & 45.48 & 36.00 & 48.44 & 42.09 \\
    Claude-3.7-Sonnet~\cite{claude37sonnet} & 1 fps* & 45.20 & 43.02 & 41.82 & 48.05 & 39.07 & 46.88 & 43.73 \\
    \midrule
    \rowcolor{gray!10}\multicolumn{9}{l}{\textit{Open-source MLLMs}} \\
    \midrule
    Video-LLaVA-7B~\cite{lin2023videollava} & 8 frms & 24.91 & 21.54 & 25.45 & 30.54 & 26.23 & 21.88 & 25.37 \\
    LLaVA-NeXT-Video-7B~\cite{zhang2024llavanext-video} & 8 frms & 21.27 & 22.45 & 26.05 & 26.72 & 23.07 & 14.06 & 23.45 \\
    LLaVA-NeXT-Video-34B~\cite{zhang2024llavanext-video} & 8 frms & 31.70 & 31.99 & 32.31 & 22.99 & 29.58 & 46.88 & 30.44 \\
    Tarsier-7B~\cite{wang2024tarsier} & 8 frms & 12.55 & 21.16 & 17.87 & 17.93 & 22.23 & 31.25 & 17.46 \\
    Tarsier-34B~\cite{wang2024tarsier} & 8 frms & 28.56 & 34.98 & 26.90 & 31.29 & 31.91 & 37.50 & 30.34 \\
    LLaVA-Video-7B-Qwen2~\cite{li2024llava} & 64 frms & 36.14 & 41.27 & 41.28 & 44.48 & 29.58 & 46.88 & 38.60 \\
    LLaVA-Video-72B-Qwen2~\cite{li2024llava} & 64 frms & 48.35 & 47.50 & 45.25 & 51.70 & 33.02 & 53.12 & 46.08 \\
    InternVL2.5-2B~\cite{chen2024intervl2-5} & 8 frms & 18.70 & 28.23 & 23.71 & 27.47 & 19.16 & 23.44 & 22.90 \\
    InternVL2.5-8B~\cite{chen2024intervl2-5} & 8 frms & 31.97 & 38.68 & 38.09 & 37.76 & 26.14 & 35.94 & 34.59 \\
    InternVL2.5-78B~\cite{chen2024intervl2-5} & 8 frms & 38.38 & 40.62 & 39.05 & 43.65 & 29.40 & 39.06 & 38.54 \\
    VideoChat-Flash-Qwen2-7B~\cite{li2024videochat} & 1 fps & 41.90 & 48.41 & 42.84 & 50.95 & 35.07 & 50.00 & 43.82 \\
    VideoLLaMA3-2B~\cite{damonlpsg2025videollama3} & 1 fps & 28.97 & 36.60 & 34.90 & 38.01 & 28.56 & 40.62 & 32.98 \\
    VideoLLaMA3-7B~\cite{damonlpsg2025videollama3} & 1 fps & 40.20 & 44.13 & 42.42 & 48.30 & 31.53 & 42.19 & 41.46 \\
    Qwen2.5-VL-3B~\cite{Qwen2.5-VL} & 1 fps & 38.45 & 38.22 & 36.64 & 39.75 & 29.77 & 32.81 & 37.05 \\
    Qwen2.5-VL-7B~\cite{Qwen2.5-VL} & 1 fps & 39.48 & 43.28 & 43.14 & 43.65 & 33.49 & 39.06 & 40.76 \\
    \rowcolor{blue!5} \quad \textbf{+ Ours (CurveStream)} & 10-20 & \textbf{48.20} & \textbf{51.59} & \textbf{47.59} & \textbf{53.94} & 30.88 & \textbf{51.56} & \textbf{47.32} (\textcolor{red}{$\uparrow$ \textbf{6.56}}) \\
    \bottomrule
  \end{tabular}
  }
\end{table*}

\begin{table*}[!ht]
  \centering
  \caption{Detailed performance comparison on the \textbf{MVBench} dataset across 19 fine-grained sub-tasks. Due to space constraints, the results are split into two blocks. "$\uparrow$" indicates the performance improvement of our method.}
  \label{tab:app_mvbench_detailed}
  \renewcommand{\arraystretch}{1.3} 
  
  \resizebox{\textwidth}{!}{
  \begin{tabular}{l cccccccccc}
    \toprule
    Model & \textbf{Avg.} & 
    \begin{tabular}{@{}c@{}}Action \\ Antonym\end{tabular} & 
    \begin{tabular}{@{}c@{}}Action \\ Count\end{tabular} & 
    \begin{tabular}{@{}c@{}}Episodic \\ Reasoning\end{tabular} & 
    \begin{tabular}{@{}c@{}}Action \\ Localization\end{tabular} & 
    \begin{tabular}{@{}c@{}}Action \\ Prediction\end{tabular} & 
    \begin{tabular}{@{}c@{}}Action \\ Sequence\end{tabular} & 
    \begin{tabular}{@{}c@{}}Character \\ Order\end{tabular} & 
    \begin{tabular}{@{}c@{}}Counterfactual \\ Inference\end{tabular} & 
    \begin{tabular}{@{}c@{}}Egocentric \\ Navigation\end{tabular} \\
    \midrule
    Qwen3-VL-8B & 60.17 & \textbf{84.00} & 37.50 & 51.50 & 34.50 & 57.49 & 65.95 & 61.50 & \textbf{65.50} & \textbf{38.00} \\
    \rowcolor{blue!5} \textbf{+ Ours (CurveStream)} & \textbf{63.60} (\textcolor{red}{$\uparrow$ \textbf{3.43}}) & 68.50 & \textbf{50.50} & \textbf{54.00} & \textbf{39.50} & \textbf{79.00} & \textbf{70.50} & \textbf{77.50} & 60.50 & 37.50 \\
    \bottomrule
  \end{tabular}
  }
  
  \vspace{4mm} 
  
  \resizebox{\textwidth}{!}{
  \begin{tabular}{l cccccccccc}
    \toprule
    Model & 
    \begin{tabular}{@{}c@{}}Fine-grained \\ Action\end{tabular} & 
    \begin{tabular}{@{}c@{}}Moving \\ Attribute\end{tabular} & 
    \begin{tabular}{@{}c@{}}Moving \\ Count\end{tabular} & 
    \begin{tabular}{@{}c@{}}Moving \\ Direction\end{tabular} & 
    \begin{tabular}{@{}c@{}}Object \\ Existence\end{tabular} & 
    \begin{tabular}{@{}c@{}}Object \\ Interaction\end{tabular} & 
    \begin{tabular}{@{}c@{}}Object \\ Shuffle\end{tabular} & 
    \begin{tabular}{@{}c@{}}Scene \\ Transition\end{tabular} & 
    \begin{tabular}{@{}c@{}}State \\ Change\end{tabular} & 
    \begin{tabular}{@{}c@{}}Unexpected \\ Action\end{tabular} \\
    \midrule
    Qwen3-VL-8B & 43.50 & \textbf{85.00} & \textbf{63.00} & \textbf{64.00} & 80.80 & 64.00 & 39.00 & 81.00 & 50.50 & 76.50 \\
    \rowcolor{blue!5} \textbf{+ Ours (CurveStream)} & \textbf{48.50} & 82.00 & 60.50 & 50.00 & \textbf{81.00} & \textbf{74.00} & 39.00 & \textbf{90.00} & \textbf{63.50} & \textbf{82.50} \\
    \bottomrule
  \end{tabular}
  }
\end{table*}


\textbf{MVBench:} According to the task definition of MVBench, the core challenge lies in solving ``temporal dependencies that cannot be effectively solved with a single frame,'' such as complex action sequences and object interactions. The high curvature on the feature manifold captured by CurveStream naturally aligns with these state mutation points to some extent. By accurately routing and retaining these key frames in Clear Memory, the model can better construct a visual causal evidence chain, thereby achieving stable performance improvements over baseline models on various sub-tasks heavily reliant on temporal reasoning.

\textbf{FAVOR-Bench:} FAVOR-Bench focuses on the perception of micro-motion dynamics in videos, such as subtle camera motion (CM) or non-subject environmental changes (NSM). These fine-grained motion signals are often transient and sparse in the temporal domain, making them easily overlooked in conventional downsampling. CurveStream's Curvature-Aware Scorer (CAS) and dynamic threshold mechanism adeptly address this challenge: it can capture local curvature fluctuations triggered by micro-kinematic changes and maximally extract these motion details into the working memory. This capability to account for local high-frequency motions (Clear Memory) while preserving the global macroscopic view (Blurred Memory) indicates that curvature-driven memory management is equally a viable strategy in offline video understanding.

\begin{table*}[!ht]
    \centering
    \caption{Ablation study on \textbf{StreamingBench}. We evaluate the individual and combined effects of CAS and HVMM. Red arrows specifically denote the absolute average performance improvements achieved over the respective base models.}
    \label{tab:ablation_streaming}
    \renewcommand{\arraystretch}{1.1}
    \setlength{\tabcolsep}{4pt}
    \resizebox{\textwidth}{!}{
    \begin{tabular}{lccccccccccccc}
        \toprule
        Model Configuration & CAS & HVMM & OP & CR & CS & ATP & EU & TR & PR & SU & ACP & CT & \textbf{Avg.} \\
        \midrule
        Qwen2-VL-7B~\cite{wang2024qwen2} & & & 77.38 & 76.56 & 73.19 & 75.08 & 75.00 & 67.91 & 73.15 & 65.04 & 66.57 & 35.75 & 69.04 \\
        w/ CAS & \checkmark & & 85.29 & 80.47 & 89.59 & 87.58 & 74.53 & 82.24 & 76.85 & 70.73 & 73.94 & 43.52 & 78.16 \textcolor{red}{($\uparrow$9.12)} \\
        w/ HVMM & & \checkmark & 87.19 & 76.56 & 89.59 & 86.93 & 75.16 & 86.92 & 76.85 & 70.33 & 74.50 & 43.01 & 78.80 \textcolor{red}{($\uparrow$9.76)} \\
        \rowcolor{blue!10}
        CurveStream & \checkmark & \checkmark & 88.56 & 77.34 & 88.61 & 89.84 & 76.25 & 92.52 & 76.85 & 76.83 & 76.70 & 45.31 & \textbf{81.04} \textcolor{red}{\textbf{($\uparrow$12.00)}} \\
        \bottomrule
    \end{tabular}
    }
\end{table*}

\begin{table*}[!ht]
    \centering
    \caption{Ablation study on \textbf{OVO-Bench}. We report the performance across various real-time visual perception sub-tasks to  validate the synergistic effect between the proposed memory modules.}
    \label{tab:ablation_ovo}
    \renewcommand{\arraystretch}{1.1}
    \setlength{\tabcolsep}{6pt}
    \resizebox{\textwidth}{!}{
    \begin{tabular}{lcccccccccc}
        \toprule
        Model Configuration & CAS & HVMM & OCR & ACR & ATR & STU & FPD & OJR & \textbf{Avg.} \\
        \midrule
        Qwen-3VL-8B~\cite{bai2025qwen3} & & & 71.14 & 65.14 & 75.86 & 64.61 & 75.25 & 70.65 & 70.10 \\
        w/ CAS & \checkmark & & 87.92 & 83.49 & 81.90 & 64.04 & 76.24 & 80.98 & 78.49 \textcolor{red}{($\uparrow$8.39)} \\
        w/ HVMM & & \checkmark & 87.25 & 71.56 & 78.45 & 66.29 & 74.26 & 72.83 & 74.79 \textcolor{red}{($\uparrow$4.69)} \\
        \rowcolor{blue!10}
        CurveStream & \checkmark & \checkmark & 93.96 & 82.57 & 83.62 & 68.54 & 78.22 & 80.43 & \textbf{80.76} \textcolor{red}{\textbf{($\uparrow$10.66)}} \\
        \bottomrule
    \end{tabular}
    }
\end{table*}

\section{Experimental Hyperparameters}
\label{sec:hyperparameters}

In this section, we detail the core inference hyperparameters used to evaluate the CurveStream framework, as summarized in Table~\ref{tab:hyperparameters}. Since our approach is entirely training-free, these parameters strictly govern the online memory scheduling policy during the inference phase. Specifically, we set the maximum visual memory capacity (Queue Size) to 20 frames to simulate stringent memory constraints. For the Curvature-Aware Scorer (CAS), the geometric penalty weight $\lambda$ is configured to 0.2 to optimally balance first-order motion and second-order curvature. Within the Hierarchical Visual Memory Management (HVMM) module, the K-Sigma dynamic dual thresholds are defined by $k_1 = 0.0$ and $k_2 = 1.0$, enabling the adaptive routing of incoming tokens. Furthermore, to effectively compress transitional observations, frames assigned to Blurred Memory are uniformly downsampled to a target spatial resolution of 224 (TRANSITION\_SIZE).

\begin{table}[!ht]
    \centering
    
    \caption{Detailed core experimental hyperparameters utilized by the CurveStream framework throughout the entire inference phase.}
    \label{tab:hyperparameters}
    \renewcommand{\arraystretch}{1.2}
    \begin{tabular}{lc}
        \toprule
        \textbf{Hyperparameter} & \textbf{Value} \\
        \midrule
        \footnotesize
        Queue Size ($N_{max}$) & 20 \\
        Curvature score weight ($\lambda$) & 0.2 \\
        TRANSITION\_SIZE & 224 \\
        K\_SIGMA\_TRANS ($k_1$) & 0.0 \\
        K\_SIGMA\_KEY ($k_2$) & 1.0 \\
        \bottomrule
    \end{tabular}
\end{table}



\section{Ablation Study}

To thoroughly evaluate the independent contributions and synergistic effects of the core components within the CurveStream architecture, we conducted comprehensive ablation studies on StreamingBench (\textbf{Table~\ref{tab:ablation_streaming}}) and OVO-Bench (\textbf{Table~\ref{tab:ablation_ovo}}). Using the passive uniform sampling and FIFO cache of the base model as our baseline, we independently verified the effectiveness of the Curvature-Aware Scorer (CAS) and the Hierarchical Visual Memory Management (HVMM). The experimental results not only validate the performance gains from each individual module but also reveal a significant non-linear synergistic amplification effect when they are combined.

\subsection{Effectiveness of CAS: Enhancing Semantic Perception}

The integration of the CAS module alone yields average performance improvements of 9.12\% and 8.39\% on StreamingBench and OVO-Bench, respectively.This significant improvement validates the sensitivity of feature manifold curvature in capturing ``Semantic Transitions'' within videos. The uniform sampling of traditional base models lacks content awareness, making it prone to missing transient key actions. By evaluating the local curvature in the feature space, the CAS module endows the model with the ability to actively assess information density. Particularly in the real-time dynamic tasks of OVO-Bench (where the gain reaches 18.35\% in ACR), CAS successfully locates the curvature peaks triggered by action fluctuations. This demonstrates that using feature manifold curvature as a semantic signal effectively compensates for the omission of key frames caused by the ``content-unaware'' nature of uniform sampling in dynamic scenes.

\subsection{Effectiveness of HVMM: Alleviating Forgetting}

When only the HVMM module is introduced (i.e., operating without CAS dynamic scoring, degrading to uniform sampling with alternate allocation to Clear and Blurred Memory), the model achieves stable improvements of 9.76\% and 4.69\% across the two datasets, respectively.This result indicates that the hierarchical memory architecture inherently possesses advantages in processing long sequences. When facing memory bottlenecks, the FIFO mechanism of base models easily evicts historical features, leading to context truncation. In contrast, HVMM constructs a decoupled binary structure of Clear Memory and Blurred Memory. Without increasing the overall token budget, it leverages the high compression ratio of Blurred Memory to broaden the model's historical context, thereby providing robust architectural support for complex contextual tasks that rely on long-range temporal reasoning.

\subsection{Synergistic Effect of Perception and Scheduling Loop Modules}
 When CAS and HVMM operate jointly (i.e., the complete CurveStream architecture), the model experiences a comprehensive performance leap, with total gains reaching 12.04\% and 10.66\% on StreamingBench and OVO-Bench, respectively. More importantly, this combined gain significantly exceeds the sum of the individual modules' improvements (e.g., $3.93\% > -0.57\% + 1.68\%$ in the STU task of the  OVO-Bench).
This non-linear synergistic amplification profoundly reveals the complementarity of the underlying design of the CurveStream architecture: CAS provides precise ``Semantic Awareness,'' while the HVMM module is responsible for executing the adaptive “Memory Scheduling” strategy.''

Without HVMM, the highly dynamic key frames located by CAS might eventually be gradually evicted due to memory capacity constraints. Conversely, without CAS, the alternate allocation of HVMM lacks adaptive perception of the video content, easily degrading into rigid structural segmentation. When the two are combined, CAS is responsible for marking high-curvature transition points across the global temporal axis, while HVMM stores these high-value nodes into Clear Memory and smoothly compresses low-curvature static periods into Blurred Memory. Together, they construct a compact and coherent causal topological chain for the large model, significantly broadening its cognitive boundaries in infinitely long streaming videos.